\documentclass{IEEEtran}

\usepackage{graphicx}
\usepackage{grffile}

\usepackage{epstopdf}
\usepackage{amsmath,amssymb,booktabs,tabularx}
\usepackage[hyphens]{url}
\usepackage{hyperref}
\usepackage{mathrsfs}
\usepackage{float}
\usepackage{multirow}
\usepackage{soul}
\usepackage[table]{xcolor}
\usepackage{tikz}
\usetikzlibrary{shapes.geometric, arrows}
\usepackage{comment}
\usepackage{subcaption}
\usepackage{mwe}
\usepackage{slashbox}
\usepackage{authblk}
\usepackage{multicol}
\usepackage{import}
\usepackage{siunitx}
\usepackage{mathtools}
\usepackage{pbox}
\usepackage{booktabs} 
\usepackage{rotating}
\usepackage{microtype}

\usepackage{makecell} 

\newcommand{\etal}[1]{#1~\emph{et~al.}}
\newcommand{\ie}{i.e.~}

\title{Bayesian Convolutional Neural Networks for Limited Data Hyperspectral Remote Sensing Image Classification}

\author{Mohammad Joshaghani, Amirabbas Davari, Faezeh Nejati Hatamian, Andreas Maier, Christian Riess
\thanks{ M.\ Joshaghani, A.\ Davari, F.\ Nejati Hatamian, A.\ Maier and C.\ Riess are with the Computer Science department at Friedrich-Alexander University Erlangen-Nürnberg, 91058 Erlangen, Germany (email: amir.davari@fau.de).}}

\begin{document}
\maketitle
\begin{abstract}
Employing deep neural networks for Hyper-spectral remote sensing (HSRS) image
classification is a challenging task. HSRS images have high dimensionality and
a large number of channels with substantial redundancy between channels. In
addition, the training data for classifying HSRS images is limited and the
amount of available training data is much smaller compared to other
classification tasks. These factors complicate the training process of deep
neural networks with many parameters and cause them to not perform well even
compared to conventional models. Moreover, convolutional neural networks
produce over-confident predictions, which is highly undesirable considering the
aforementioned problem. 

In this work, we use for HSRS image classification a special class of deep
neural networks, namely a Bayesian neural network (BNN). To the extent of our
knowledge, this is the first time that BNNs are used in HSRS image
classification. BNNs inherently provide a measure for uncertainty. We perform
extensive experiments on the Pavia Centre, Salinas, and Botswana datasets. We
show that a BNN outperforms a standard convolutional neural network (CNN) and
an off-the-shelf Random Forest (RF). Further experiments underline that the BNN
is more stable and robust to model pruning, and that the uncertainty is higher
for samples with higher expected prediction error.

\end{abstract}


\section{Introduction}\label{sec:intro}

Hyperspectral remote sensing (HSRS) images serve many applications, ranging
from urban planning, agricultural region monitoring, and natural resource
management to material identification~\cite{hsrs_app1}. 
HSRS images are such a rich sources of information due to their dozens or
hundreds of channels of different spectral wave-lengths~\cite{hsrs_app2}. 
On one
hand, this high dimensionality demands a large number of training data. On the
other hand, training data is typically expensive to acquire.
These two factors and the fact that the training data does not scale with the data dimensionality result in the Hughes phenomenon~\cite{hughes}.
This phenomenon challenges the training process of parametric machine
learning methods, leading to over- and underfitting, inaccuracy, and
convergence to non-optimal solutions.

This issue becomes more pressing with the increasing popularity of
convolutional neural networks (CNNs) for classification tasks.
CNNs reach outstanding performance on many computer vision tasks such as image
classification.
Deeper CNNs are capable of extracting more expressive features and modeling
more complex functions. However, the number of parameters scales with the depth
of the network. Hence, limited training data easily cause the CNN to overfit.

In this work, we explore Bayesian neural networks (BNN) to alleviate this
challenge. BNNs are a specific variant of deep neural networks.  In BNNs, the
network parameters of one or more layers are replaced by probability
distributions. When performing network inference, a specific set of weights is
sampled from these distributions. Hence, one can view a BNN as a distribution
of networks. This distribution enables ensembling by performing multiple
inference steps. Then, the prediction is the mean of the ensemble results, and
their variance can serve as an additional uncertainty measure.

Ensembling is a well-known and effective tool to tackle overfitting, and we
show in our experiments that BNNs indeed improve the overall classification
performance. Additionally, the learned weight distributions of the BNN can
provide valuable insights. We specifically use these weight distributions to
prune the network by removing the least informative weights from the network. 

In detail, our specific contributions are
\begin{enumerate}
\item We propose Bayesian Neural Networks for HSRS image classification on
limited training data. We demonstrate its performance in comparison to a
comparably constructed standard CNN and a classical random forest classifier.
To our knowledge, this is the first time that BNNs are used for HSRS image
classification.
\item We show two applications of the network. First, we demonstrate that the
distributions of weights in the BNN can be effectively used to prune the
network, which makes the BNN more compact and
enables its use also on resource-limited devices. This pruning reduces the
original parameter set by an order of magnitude with only small impact on
classification performance. Second, we gradually omit test samples with high
uncertainty to show that the samples with uncertainty are indeed more likely to
be misclassified.
\end{enumerate}

The rest of this paper is organized as follows: Section~\ref{sec:related_work} reviews the related work. Section~\ref{sec:background} briefly overviews the mathematical background of the Bayesian neural network. Section~\ref{sec:Methodology} explains the proposed HSRS classification pipeline. In Sec.~\ref{sec:exps_train_simple}, the experiments and their settings, as well as their results are discussed. Finally, Sec.~\ref{conclusion} concludes this work.

\section{Related Works}\label{sec:related_work}
This section is organized in three parts. We first discuss works that address
the issue of limited  data in HSRS image classification. Then, we review
convolutional neural networks for HSRS image classification. Third, we
introduce prior work on Bayesian neural networks.

\textbf{Limited Training Data.} Several works have been conducted to tackle the issue of limited training data. As suggested by \etal{Davari}~\cite{fast_and_efficient}, the solutions can be classified in three classes: 1) Developing new architectures or adapt techniques in the current architectures, such as regularizers and data augmentation, to boost network's performance in this condition \cite{lr_lim10, lr_lim11, lr_lim12}, 2) Reducing the dimensionality of the feature vectors to feed the classifiers more informative data, and 3) Artificially generating synthetic data to increasing the number of training data \cite{fast_and_efficient, davari2018gmm, davari2015effect}. 

Since there is a large redundancy in spectral channels in HSRS images, applying dimensionality reduction techniques, such as PCA, is highly effective. In \cite{lr_lim20}, to reduce the dimensionality of edge preserving filters, PCA is used. The dimensionality of the filters are reduced before feeding them to the classifier. The authors show that such feature vectors are powerful and provide performance improvement. Other works use PCA after the feature vector, e.g., extended multi-attribute profile (EMAP)~\cite{dalla2010morphological}, to provide a more powerful and informative feature vector to the model~\cite{fast_and_efficient, davari2018gmm}. Following them, we also adapt this approach and use PCA on EMAP features to improve the performance.

Although promising, synthetically populating severely small training data using generative models is limited. Several works use GANs \cite{gan} to generate the synthetic data to be added to the training pool~\cite{gan1, dietrich2021synthetic}. The main disadvantage of these methods is the fact that generative models such as GANs require a relatively high amount of training data in order to capture the underlying distribution of the data accurately. 


\textbf{CNNs in HSRS image classification.} Deep CNN networks are dominant in many computer vision tasks, such as classification.
Recently, CNNs also started to receive attention from researchers in HSRS image classification. They are reported to outperform established HSRS classification baselines.
Additionally, it has been reported that using spatial and spectral features
together will increase the classification accuracy \cite{rw_spatial1,
rw_spatial2}, which is one advantage of convolutional-based architectures used
on HSRS images is they inherently use data across the channel as well.

\etal{Chen}~\cite{chen2014deep} proposed stacked autoencoders (SAE) to learn abstract HSRS image features in an unsupervised fashion.
The resulting feature set was then fed into a logistic regression for classification of HSRS images.
In another work, the same authors proposed to replace SAE by a deep belief network (DBN).
Considering the similarity of DBNs and CNNs, their work is one of the first ones that pioneered the use of deep neural network for HSRS image classification.
\etal{Makantasis}~\cite{deep1} used a 2-D CNN architecture, and reduces the dimensionality of the data via PCA to three dimensions, trying to process it as RGB image data.
However, some other works use 3-D CNN that work with raw data or with dimensionality-reduced data~\cite{deep2, deep3}. \etal{Audebert} provides a review article deep learning methods that are used in HSRS image classification~\cite{deep_lr}.

It is a difficult task to perform CNN training for HSRS image classification with only very limited training data.
\etal{Chen}~\cite{chen2017hyperspectral} proposed to combine Gabor filters with convolutional filters, where the Gabor filters are used to encode spatial information like textures and edges.
Other works explore various data augmentation algorithms to mitigate the limited training data size~\cite{aptoula2016deep, li2018data}.
A general strategy is to fine-tune a network that was pre-trained on a larger datasets like ImageNet. Eventhough the data for fine-tuning itself is insufficient, this training strategy generally improves the classification performance. \etal{He}~\cite{he2020transferring} proposed transfer learning in combination with ensemble learning. They used a ImageNet-pretrained network and randomly selected three channels of the HSRS image to fine-tune multiple networks.
Finally, they used the resulting networks in an ensemble learning scheme.
Alternatively, active learning has been utilized to remedy the limited data problem. \etal{Cao}~\cite{cao2020hyperspectral} proposed to iteratively train a CNN on a small training set. Then, they select the most informative pixels from a candidate pool, and fine-tune the network on the new training dataset.

\textbf{Bayesian neural networks.} Several studies have applied Bayesian
methods to neural networks. The main challenge is that it is intractable to
compute the true posterior probability distribution. Therefore, different
approximation methods have been investigated and proposed to compute the
posterior. Buntine and Weigend~\cite{b1} proposed different maximum
a-posteriori (MAP) schemes for neural network and considered second order
derivatives in the prior probability approximation. Several other attempts have
been made to improve the approximation quality while keeping the computation
tractable and applicable on modern applications~\cite{blundell, b4}. However,
two approaches are the most successful among them. The first one is to
construct the Bayesian network using Dropout \cite{drop} and Gaussian Dropout
\cite{gdrop} as approximate variational inference schemes \cite{bcnn_gubin}.
The second approach builds the Bayesian neural network using Backprop
\cite{bayes_by_backprop1, blundell} based on variational inference. It
considers Gaussian probability distribution over the weights, which are defined
using two parameters, mean and variance \cite{softplus}. Both approaches
provide methods to approximate uncertainty. In \cite{softplus}, a comparison of
the two approaches is made and they are shown to perform comparably on MNIST
dataset \cite{mnist}. As explained in \cite{softplus}, the Softplus
normalization, applied to \emph{Bayes by Backprop}-based variational inference
to estimate uncertainties, is more appropriate to use in computer vision tasks. 

To the extent of our knowledge, Bayesian neural networks have never been
employed in this field. In this work, our focus is to employ them on HSRS image
classification task. We perform extensive evaluation of the Bayesian model and
compare different aspects of the model with a conventional (non-Bayesian) model.

\begin{figure*}[tb]
	\centerline{\includegraphics[width=.8\linewidth]{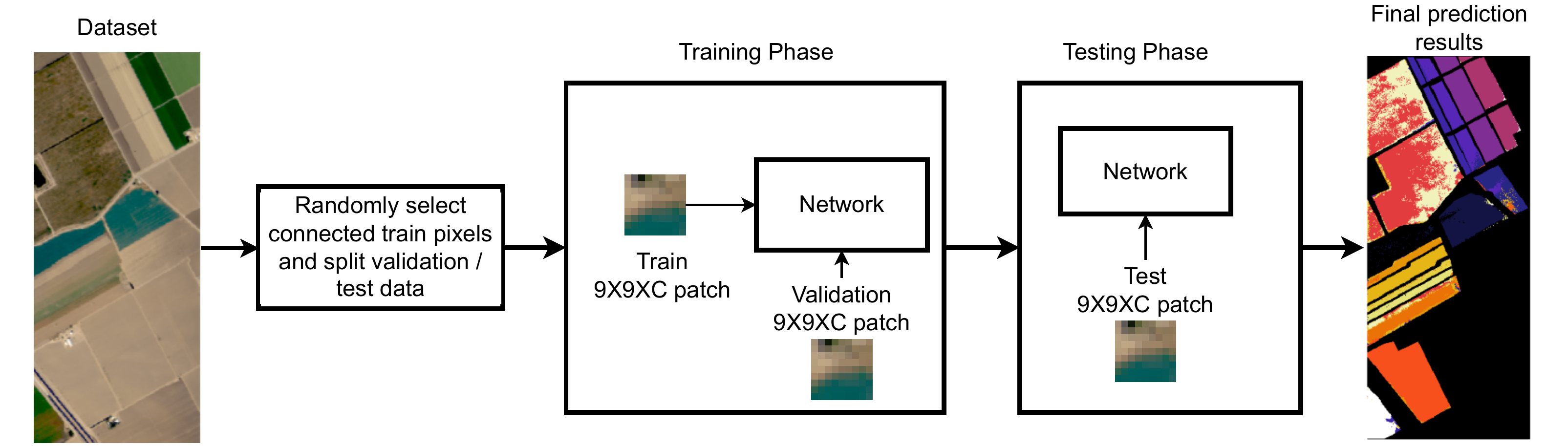}}
	\caption{HSRS image classification pipeline.}
	\label{fig:clasificaiton_pipeline}	
\end{figure*}

\section{Bayesian Convolutional Neural Network}\label{sec:background}
Frequentist neural networks have a tendency to produce overconfident
predictions.  Besides, when not provided with enough training data, they are
prone to the overfitting problem. Combining the idea of Bayesian learning with
standard neural networks can address these limitations.
Bayesian models perform predictions by integrating over the distribution of possible models and the prior probability. This enables an intrinsic regularization making them more robust against overfitting.

The BNN in this work follows the work by \etal{Shridhar}~\cite{shridhar1}. It
is a Bayesian Convolutional Neural
network (BCNN) that uses Variational Inference. This section provides a brief
background on the network. For a more in-depth discussion, the reader is
referred to~\cite{shridhar1}.

\subsection{Variational Inference}
Bayesian neural networks train a model by inferring the model posterior.
Accurate inference of the model posterior is computationally demanding, and
even for only moderately sized models intractable. Hence, the model posterior
is usually approximated. One popular and successful approximation method is
variational inference. 

Given the input set $\mathbf{X}={\mathbf{x}_1, \mathbf{x}_2, ..., \mathbf{x}_N}$ and a corresponding output set $\mathbf{y}={y_1, y_2, ..., y_N}$, then the functor $f(\mathbf{X}) = y$ estimates the output $y$ from the inputs $\mathbf{X}$.
Bayesian learning provides a principled approach to obtain the model posterior
$p(f | \mathbf{X}, \mathbf{y})$. Two components are required to calculate the posterior. First, 
a prior distribution $p(f)$ that captures a prior belief about the estimator functions.
Second, a likelihood function $p(\mathbf{y}|f,\mathbf{X})$ for indicating how
likely it is for the model $f$ to predict the output $y$ given the observations $\mathbf{X}$.

More specifically, given an unseen data $(\mathbf{x}^*,y^*)$, the posterior is
obtained by integrating over all possible estimator functions $f$ that are
parametric models with parameter set $\mathbf{\theta}$,
\begin{equation} \label{eq:bayes_inference}
	\begin{split}
		p(y^*|\mathbf{x}^*,\mathbf{X},\mathbf{y})=\int p(y^*|f)p(f|\mathbf{x}^*,\mathbf{X},\mathbf{y})df \\
		=\int p(y^*|f)p(f|\mathbf{x}^*,\mathbf{\theta})p(\mathbf{\theta}|\mathbf{X},\mathbf{y})dfd\mathbf{\theta}\enspace.
	\end{split}
\end{equation}
This integral is intractable due to the intractability of the distribution $p(\mathbf{\theta}|\mathbf{X},\mathbf{y})$. Hence, the variational approach is to approximate $p(\mathbf{\theta}|\mathbf{X},\mathbf{y})$ with a variational distribution $q(\mathbf{\theta})$.
The candidate $q(\theta)$ should be as similar as possible to the original intractable distribution. The similarity of $p(\mathbf{\theta}|\mathbf{X},\mathbf{y})$ and $q(\mathbf{\theta})$ can be assessed with the Kullback-Leiber (KL) divergence~\cite{kullback}. 
Minimizing the mentioned KL divergence is equivalent to maximizing the log evidence lower bound with respect to the parameter set $\mathbf{\theta}$
		\begin{equation}\label{eq:logkl}
			\text{KL}_{\text{VI}} = \int q(\mathbf{\theta})p(F|\mathbf{X}, \mathbf{\theta})\log_{p}(\mathbf{y}|F)dFd\mathbf{\theta} - \text{KL}(q(\mathbf{\theta})||p(\mathbf{\theta}))\enspace.
		\end{equation}
%

Maximizing $\text{KL}_{\text{VI}}$ results in a variational function that approximates the posterior.
The approximation $q(\mathbf{\theta})$ simplifies equation \ref{eq:bayes_inference} to

\begin{equation}\label{eq:bayes_approximation}
	q(y^*|\mathbf{x}^*)= \int p(y^*|f)p(f|\mathbf{x}^*,\mathbf{\theta})q(\mathbf{\theta})dfd\mathbf{\theta}\enspace.
\end{equation}
When the network performs inference, the network parameters $\mathbf{\theta}$ are sampled from $q(\mathbf{\theta})$. 

\subsection{BNN Training via Back-Propagation}
In \emph{Bayes by Backprop}~\cite{blundell, bayes_by_backprop1}, the posterior distribution on the neural network's weights is learned. Since the true posterior is often intractable, an approximate distribution $q_{\alpha}(\mathbf{\theta})$ similar to the true distribution $p(\mathbf{\theta})$ is defined.
The training consists of minimizing the KL divergence of $q_{\alpha}(\mathbf{\theta})$ and the true intractable posterior $p(\mathbf{\theta})$ by finding the optimal parameter $\alpha$. This is done by approximating the integral from Eq.~\ref{eq:logkl} with $n$ drawn samples, 
%
\begin{equation}\label{eq:cost_function}
	F(\mathcal{D}, \alpha) \approx \sum_{i=1}^{n}\log{q_{\alpha}}(\mathbf{\theta}^{(i)}|\mathcal{D}) - \log{p}(\mathbf{\theta}^{(i)}) - \log{p}(\mathcal{D}|\mathbf{\theta}^{(i)})\enspace,
\end{equation}
where $\mathcal{D}$ is the training dataset. $\mathbf{\theta}^{(i)}$ is a sample from the variational distribution $q_\alpha(\mathbf{\theta}|\mathcal{D})$, which we set as a Gaussian distribution
with mean and standard deviation as parameters.
The cost function in Eq.~\ref{eq:cost_function} consists of three terms. 
First, $\log{q_{\alpha}}(\mathbf{\theta}^{(i)}|\mathcal{D})$
is the variational posterior with mean $\mu$ and standard deviation $\sigma$,
\begin{equation}\label{eq:cost_term1}
	\log{q_{\alpha}}(\mathbf{\theta}^{(i)}|\mathcal{D}) = \sum_{i}^{}\log\mathcal{N}(\mathbf{\theta}_i|\mu,\sigma^2)\enspace.
\end{equation}
Second, $\log{p}(\mathbf{\theta}^{(i)})$ denotes the log prior, which is a zero-mean Gaussian distribution
		\begin{equation}\label{eq:cost_term2}
			\log{p}(\mathbf{\theta}^{(i)}) = \sum_{i}^{}\log{\mathcal{N}}(\mathbf{\theta}_i|0,\sigma_p^2)\enspace.
		\end{equation}
Third, the likelihood $\log{p}(\mathcal{D}|\mathbf{\theta}^{(i)})$ is the network output.
%
%
%

\subsection{Bayesian Neural Network with Variational Inference}
To fuse Bayesian learning in CNNs and tackle the intractable posterior distribution problem via variational inference, it is necessary to build convolutional layers with a probability distribution over the weights as filter weights, as well as fully-connected layers. In this case, the weights would be samples from the corresponding distribution. As mentioned in the previous section, the distributions are Gaussian, and each weight distribution is defined by its mean $\mu$ and variance $\sigma$.

To adapt this idea on the convolutional layers, \etal{Shridhar} used Local Re-parameterization technique \cite{variational}. It is simply re-writing and re-parameterizing the equations above to translate the global uncertainty to the local uncertainty. Using this trick, instead of directly sampling the weights, the activation maps $b$ are sampled, leading to more efficient and faster computational.

Consider the network weights variable $\mathbf{w}$. The variational posterior $q_\mathbf{w}(\mathbf{w}_{ijhw}|\mathcal{D})$ = $\mathcal{N}(\mu_{ijhw},\alpha\mu_{ijhw}^2)$, where $i$ and $j$ are inputs, and $h$ and $w$ are filter height and width. This results in the following for the activation of convolutional layer for the corresponding receptive field $R_i$:
\begin{equation}\label{eq:conv_act}
	b_j = R_i * \mu_i + \epsilon_j \odot \sqrt{R_i^2*(\alpha_i \odot \mu_i^2)}\enspace,
\end{equation}
where $\epsilon_j \sim \mathcal{N}(0,1) $, $\odot$ is the element-wise multiplication, and $*$ is the convolution operator.
As it can be observed, this trick transforms the convolutional operation within a layer into two operations: First, the output of $b$ is treated as a frequentist output and gets updated by Adam optimizer. This single-point estimate is considered as the mean of the posterior. In the second operation, the variance of the distribution is learned. This formulation ensures a non-zero positive variance. \etal{Shridhar} introduced Softplus activation function to accommodate both operations \cite{softplus}.

\subsection{Softplus Activation Function}
Two operations are applied in the convolutional and fully connected layers. One accounts for mean $\mu$, and the other accounts for the variance $\alpha\mu^2$. Applying Softplus \cite{softplus} ensures a non-zero positive variance after the training. It is a smooth approximation of ReLU activation function, but it does not reach zero when the input approaches minus infinity. Softplus is defined as

\begin{equation}\label{eq:softplus}
	\text{Softplus}(x)=\frac{1}{\beta}\log(1+\exp(\beta x))\enspace.
\end{equation}

\noindent By default, $\beta$ is set to $1$.

\subsection{Uncertainty in Bayesian Neural Networks}
One of the main advantages of Bayesian neural networks over frequentist neural networks is the ability to express uncertainty.
Bayesian learning allows to distinguish two types of uncertainty, namely \emph{aleatoric} and \emph{epistemic} uncertainty~\cite{aleatoric_or_epistemic}.
\emph{Aleatoric} uncertainty captures uncertainty caused by the data, e.g.,
inherent noise in data, whereas \emph{epistemic} uncertainty captures the model
uncertainty. For example, this implies that if the model is provided with more
data from the same underlying distribution, then the epistemic uncertainty will
be reduced, but the aleatoric uncertainty will remain at the same level. 


On real data, there is typically no closed-form solution for the predictive distribution. However, an unbiased estimator of the expected value of the predictive distribution can be written as
\begin{equation}\label{eq:predic_dist}
	\mathbb{E}_q(p_\mathcal{D}(y^*|\mathbf{x}^*))=\int q_\alpha(\mathbf{w}|\mathcal{D})p_\mathbf{w}(\mathbf{y}|\mathbf{X})d\mathbf{w}\\
\end{equation}
\begin{equation}\label{eq:predic_dist0}
	\approx \frac{1}{T}\sum_{t=1}^{T}p_{\mathbf{w}_t}(y^*|\mathbf{x}^*)\enspace,
\end{equation}
where $T$ is the number of samples.
The variance of this expected value can be decomposed into the aleatoric and epistemic uncertainty. 
More specifically, the \emph{aleatoric} uncertainty is calculated as
		%
		\begin{equation}\label{eq:aleatoric}
			Aleatoric=\frac{1}{T} \sum_{t=1}^{T}(\text{diag}(\hat{\mathbf{p}}_t)) - \hat{\mathbf{p}}_t\hat{\mathbf{p}}_t^\text{T} \enspace,
		\end{equation}
and the epistemic uncertainty is calculated as
		\begin{equation}\label{eq:epistemic}
			Epistemic=\frac{1}{T} \sum_{t=1}^{T}(\hat{\mathbf{p}}_t - \bar{\mathbf{p}})(\hat{\mathbf{p}}_t - \bar{\mathbf{p}})^\text{T} \enspace,
		\end{equation}
		%
%
where $\hat{\mathbf{p}}_t$ denotes the network output, \ie $\text{Softmax}(f_{\mathbf{w}_t}(\mathbf{x}^*))$, and $\bar{\mathbf{p}}$ is the average of $\hat{\mathbf{p}}_t$ over different samples, \ie $\frac{1}{T} \sum_{t=1}^{T}\hat{\mathbf{p}}_t$.
For a derivation of these equations, the reader is referred to \cite{shridhar1}.

\section{Bayesian Neural Network for HSRS image classification}\label{sec:Methodology}
Figure~\ref{fig:clasificaiton_pipeline} shows an overview of the classification pipeline.
The data is split into training, validation, and test sets. Then, the model is trained on the training set, and after each epoch, the model is validated.
After the training, the model with the best validation kappa score is selected and evaluated on the test set. 

Deep neural networks are in principle capable of extracting the features from raw input data.
However, quality and quantity of the training data are important requirements for training convergence to a good-performing feature extractor.
When the limited data is scarce, the network may not converge.
In such a case, they can benefit from a pre-processing step that eliminates redundancy and reduces the feature dimensionality.
In this spirit, we calculate EMAP features and further reduce the feature
vector via PCA analogously to previous works, e.g.,
\etal{Aptoula}~\cite{aptoula2016deep}.

We study three aspects to assess the benefits of Bayesian neural networks for the classification of limited-data HSRS images:

\subsection{Comparison of Classification Performance of BNN versus Standard CNN}
We first investigate how the BNN performance compares to the performance of a standard CNN with identical architecture. 
Furthermore, the BNN can be considered as a distribution of networks from which a (theoretically infinite) number of networks can be sampled. Ensembling generally leads to performance improvement and robustness against overfitting. Hence, we specifically add a comparison between the BNN before and after ensembling.

\subsection{Robustness Against Overfitting}
Overfitting is one of the main challenges that severely limited training data imposes to the learning process.
We monitor the evolution of the network performance on the training and validation sets. This provides an effective tool to identify the existence of overfitting.
We compare the BNN and CNN side-by-side to show that the BNN is more robust
against overfitting on limited training data.

\subsection{Model Pruning}
The distribution of the weights may provide additional insights into the network quality.
Performing the classification on a smaller sub-network may indicate that less network weights are involved in the largest contribution to the prediction performance.
We investigate this behavior by iteratively pruning the model weights for both the BNN and CNN while observing their classification performance.

\section{Experimental Results}\label{sec:exps_train_simple}

\subsection{Dataset}
\begin{figure}[tb]
	\begin{minipage}[b]{1\linewidth}
		\centering
		\centerline{\includegraphics[width=1\linewidth]{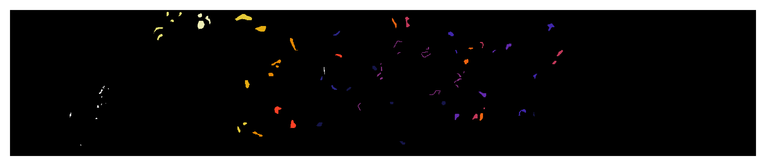}}
		\centerline{(a)}\medskip
	    \end{minipage}

		\begin{minipage}[b]{0.48\linewidth}
		\centering
		\centerline{\includegraphics[height=1.8\linewidth , width=1\linewidth]{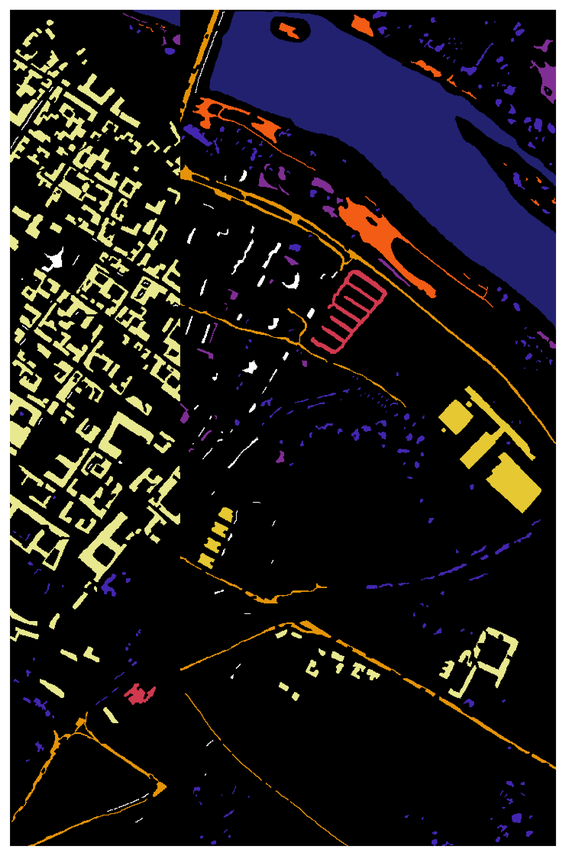}}
		\centerline{(b)}\medskip
	    \end{minipage}
	    		\begin{minipage}[b]{0.48\linewidth}
		\centering
		\centerline{\includegraphics[height=1.8\linewidth, width=1\linewidth]{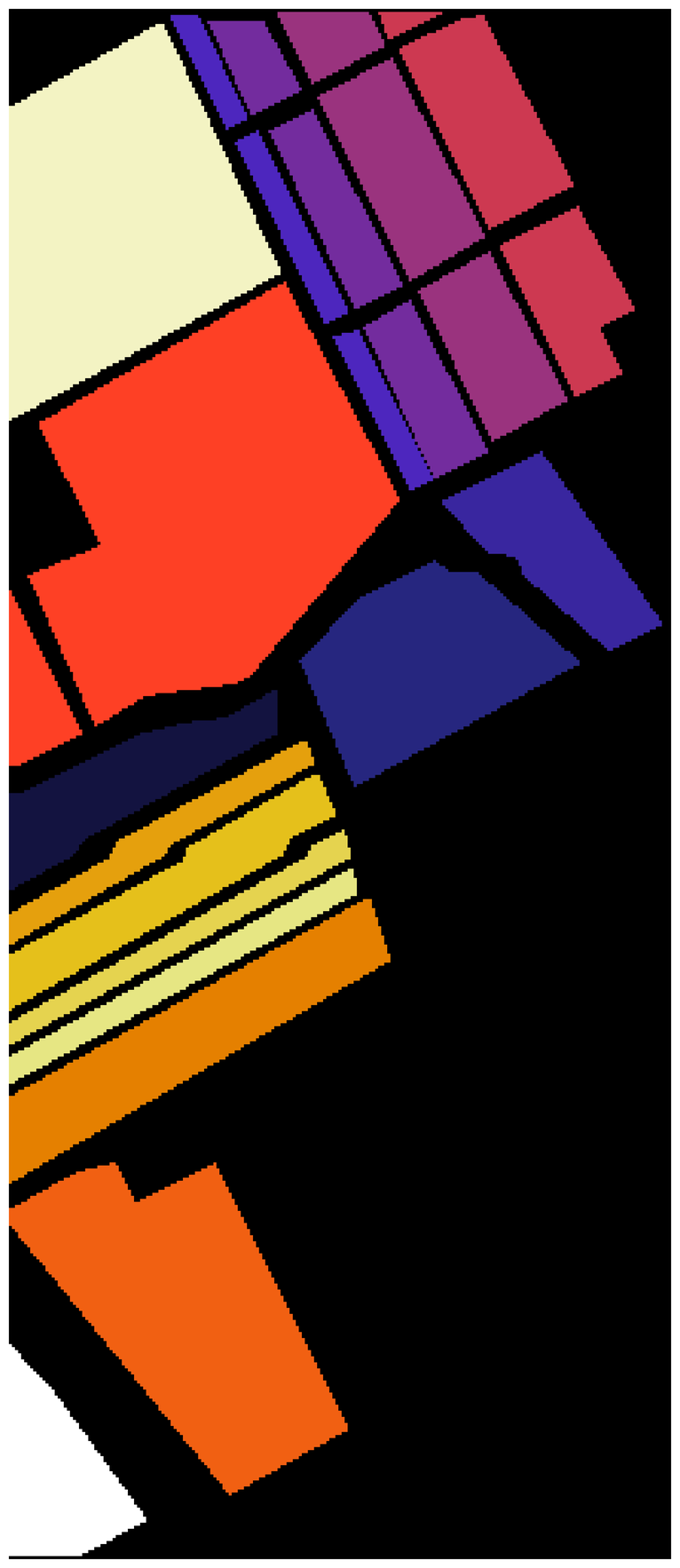}}
		\centerline{(c)}\medskip
	    \end{minipage}
	\captionof{figure}{(a) Botswana, (b) Pavia Centre, and (c) Salinas datasets ground truth classes.}
	\label{fig:all_datasets_labels}
\end{figure}

In this work, we use the commonly used and publicly available Pavia Centre, Salinas, and Botswana datasets for evaluation. Figure~\ref{fig:all_datasets_labels} illustrates the class distribution of the datasets. Salinas dataset was acquired by the AVIRIS sensor over the Salinas Valley, California. There are 512 $\times$ 217 samples in the dataset. It is comprised of 16 classes, including different types of vegetation, fields and soil.

Botswana dataset was acquired by NASA EO-1 satellite using the Hyperion sensor in 242 bands over the Okavango Delta, Botswana. After removing the noisy bands, the raw dataset has 145 spectral channels. This dataset comprises of 14 classes, including different swamps and woodlands. The number of available labeled pixels is 3248.

The Pavia Centre dataset has been acquired by the ROSIS sensor in 102 spectral bands over Pavia, northern Italy. The original scene is a 1096$\times$ 1096 image, however, some sections of the image contains no data and is blank. After removing that section, the resolution would be 1096 $\times$ 715 pixels. This dataset contains 148152 labeled pixels in 9 classes. Compared with the other two datasets, this dataset has larger homogeneous regions.

\subsection{Experimental Setup}

\subsubsection{Pre-processing}
Throughout all experiments, we use EMAP-PCA as feature descriptor~\cite{fast_and_efficient, davari2018gmm}.
This descriptor first applies PCA to the raw hyperspectral image data. The eigenvectors that are associated with the largest eigenvalues preserve 99\% of the total spectral variance, while greatly reducing redundancies between spectral bands. Then, extended multi-attribute profile (EMAP) features~\cite{dalla2010morphological, dalla2010extended} are computed. These features require four thresholds $\lambda$ for the area of connected components, which we set as $\lambda = [100, 500, 1000, 5000]$. Finally, we use PCA again to preserve 99\% of the feature variance. Before feeding the data to the network, the feature vectors go through min-max normalization, which scales them between $[0,1]$.

\subsubsection{Data Split}
In order to best evaluate the model performance, we split the data into train,
validation, and test sets. The model is trained and then evaluated, and the
model with the best evaluation scores is selected and again evaluated with the
test set. The provided results are obtained from the test set.

We devoted particular attention to the composition of the training and test
splits, in order to reduce avoidable correlations between training, validation,
and test data. More in detail, the neural network filters operate on spatial
regions of $9\times 9$ pixels centered at a pixel of interest. Hence, it is
desirable that all pixels in this $9\times 9$ region belong to the same dataset
(training, validation, test), to realistically assess the generalization
performance. To this end, the training data pixels are not simply randomly
drawn from the whole dataset. Instead, one pixel per class is drawn randomly,
and only the connected component of pixels with the same class as the randomly
drawn pixel is used as a basis for the training. More details on this approach
are presented in the supplemental material.

\subsubsection{Experimental Setup and Hyperparameters}
The experimental comparison includes two baseline methods, namely a standard feed-forward CNN with a similar architecture to the proposed BNN, and an off-the-shelf Random Forest \cite{rf_breiman}.
For brevity and disambiguation, we denote throughout our experiments the standard CNN as ``CNN'', and the proposed Bayesian Convolutional Neural Network as ``BNN''. 

The loss function for training the standard CNN is the categorical cross-entropy. The loss function for training the BNN is the Variational Gaussian loss. The random forest objective function is the Gini index. 
The optimizer is set to Adam with initial learning rate of $0.001$. Data augmentation is used to increase the variety in the training data: vertical flips, and rotations of 0, 90, 180, and 270 degrees are randomly applied to incoming patches. The model performance is measured with the overall accuracy, the average class accuracy, and the kappa score.

The architectures of the BNN and the CNN networks are identical and detailed in Tab.~\ref{tab:architecture}. It consists of three convolutional blocks and one fully connected layer.
The convolutional blocks consist of a convolutional layer, followed by an activation function, followed by layer normalization.
The activation function within the CNN blocks in the CNN is ReLU and in the BNN is Softplus.
The receptive field size in the three convolutional layers is $3 \times 3$. The number of filters in these three layers are $128$, $256$, and $512$.
The fully connected layer has $4068$ nodes. In the last layer, the activation function is Log Softmax.
To simulate the limited training data situation, $20$ pixels per class have been chosen as the training set. Since there are multiple possibilities of selecting training set, we perform the experiments $20$ times and report the mean and standard deviation of the performance metrics.

BNN inference is done by drawing 50 samples, which can be seen as an ensemble
of 50 networks. For comparison, we also report in the first experiment results
for performing only a single BNN draw, denoted as ``Bayesian NN (single
draw)''.

The first experiment investigates the classification performance and studies the robustness to overfitting.
%
The second experiment investigates the performances when progressively pruning model weights. The third experiment illustrates the uncertainty metric by the BNN and its reliability.

\begin{table}
	\centering
	\caption{Network architecture details.}
	\begin{tabular}{@{}l@{}c@{}c@{}c@{}r@{}}
		Block Name \hspace{5mm} & input size  \hspace{5mm}& output size \hspace{7mm}& kernel size \hspace{7mm}& \#filters\\
		\hline
		CNN block 1   & X$\times$9$\times$9 \hspace{5mm}  & 128$\times$7$\times$7 & 3$\times$3 & 128\\
		CNN block 2   &   128$\times$7$\times$7  \hspace{5mm}   &  256$\times$5$\times$5 & 3$\times$3 & 256 \\
		CNN block 3 	&     256$\times$5$\times$5    \hspace{5mm}    &   512$\times$3$\times$3 & 3$\times$3 & 512 \\
		\hline
		Flatten &       512$\times$3$\times$3 \hspace{5mm}     &   4068  & &\\            
		\hline
		Dense 1 &        4068 \hspace{5mm}    &  \# labels  & & \\            
		\hline
		Log Softmax &       \# labels  \hspace{5mm}    &  \# labels & &\\
		\hline
	\end{tabular}
	\label{tab:architecture}
\end{table}

\begin{table}[t]
	\centering
	\caption{Comparison of classification performance between random forests, standard CNN, and the proposed Bayesian NN with a single inference step and 50 inference steps.}
	\begin{tabular}{l@{ } c@{ } c@{ }}
		\hline
		Model& Kappa Score & Overall Acc. \\ \hline \hline

		\rowcolor{lightgray}
		\multicolumn{3}{c}{Pavia Dataset} \\ \hline
		Random Forest       & 0.7730$\pm$0.0535 & 83.64$\pm$4.02 \\ 
		CNN &0.8849$\pm$0.0300  &91.84$\pm$3.00 \\
		Bayesian NN (single draw) & 0.8943$\pm$0.0220 & 92.52$\pm$1.65 \\
		Bayesian NN & 0.9034$\pm$0.0185 & 93.19$\pm$1.29\\ \hline
		\hline

		\rowcolor{lightgray}
		\multicolumn{3}{c}{Botswana Dataset} \\ \hline
		Random Forest & 0.8376$\pm$0.0277 & 85.02$\pm$2.55\\
		CNN	& 0.8987$\pm$0.0164 & 90.65$\pm$1.52 \\
		Bayesian NN (single draw) & 0.9315$\pm$0.0128 & 93.68$\pm$1.18\\
		Bayesian NN & 0.9383$\pm$0.0137 & 94.31$\pm$1.26\\ \hline 
		\hline 
		
		\rowcolor{lightgray}
		\multicolumn{3}{c}{Salinas Dataset} \\ \hline
		
		Random Forest & 0.7481$\pm$0.0318    & 77.23$\pm$2.94     \\
		CNN  & 0.7857$\pm$0.0197   & 80.82$\pm$1.74     \\        	
		Bayesian NN (single draw) & 0.7870$\pm$0.0278   & 80.77$\pm$2.46     \\
		Bayesian NN & 0.7886$\pm$0.0277 & 80.92$\pm$2.48\\ \hline
		
	\end{tabular}%
	\label{tab:trainresultsall}%
\end{table}%


\begin{figure}[t]
	\includegraphics[width=0.48\linewidth]{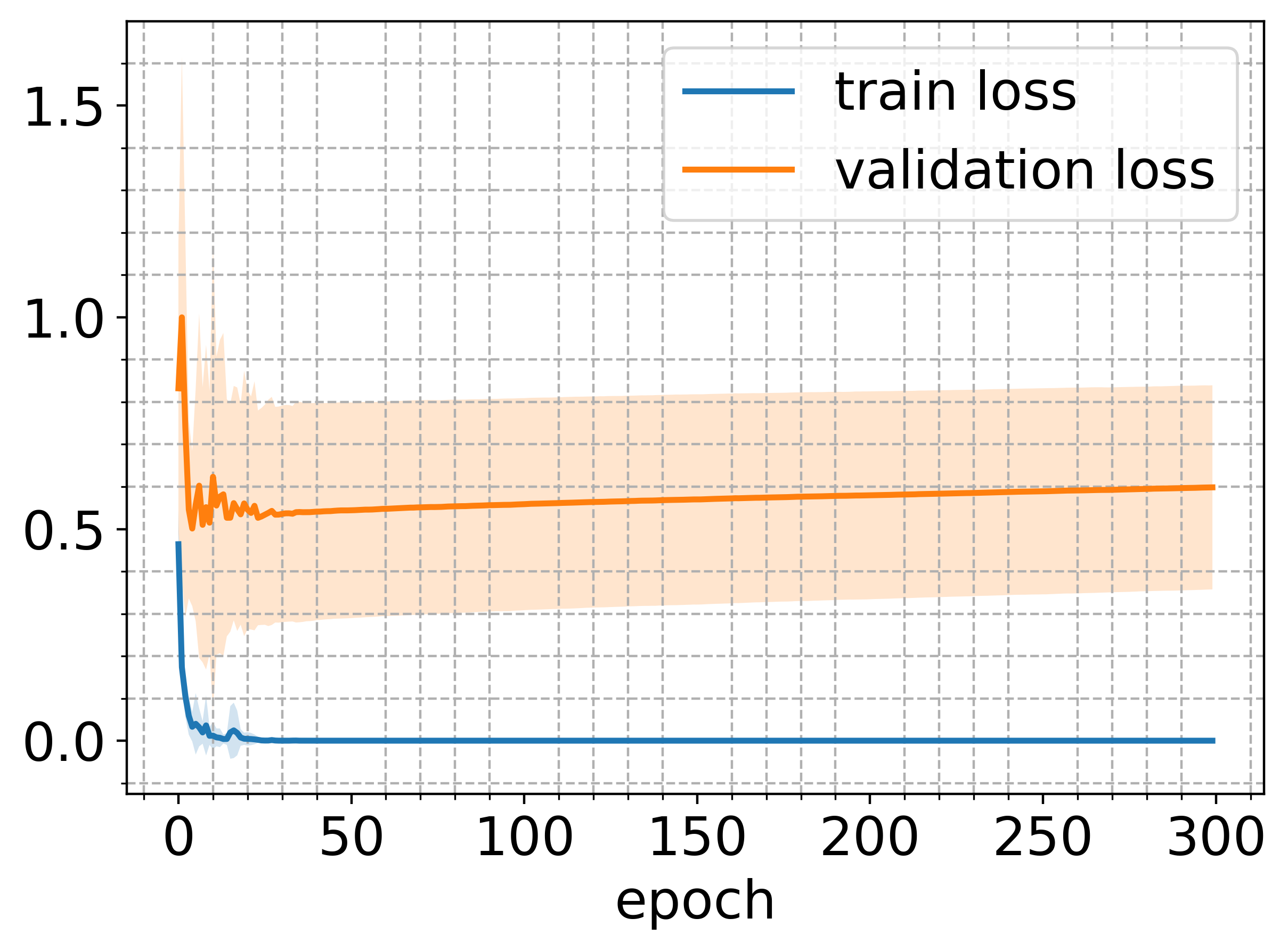}
	\includegraphics[width=0.48\linewidth]{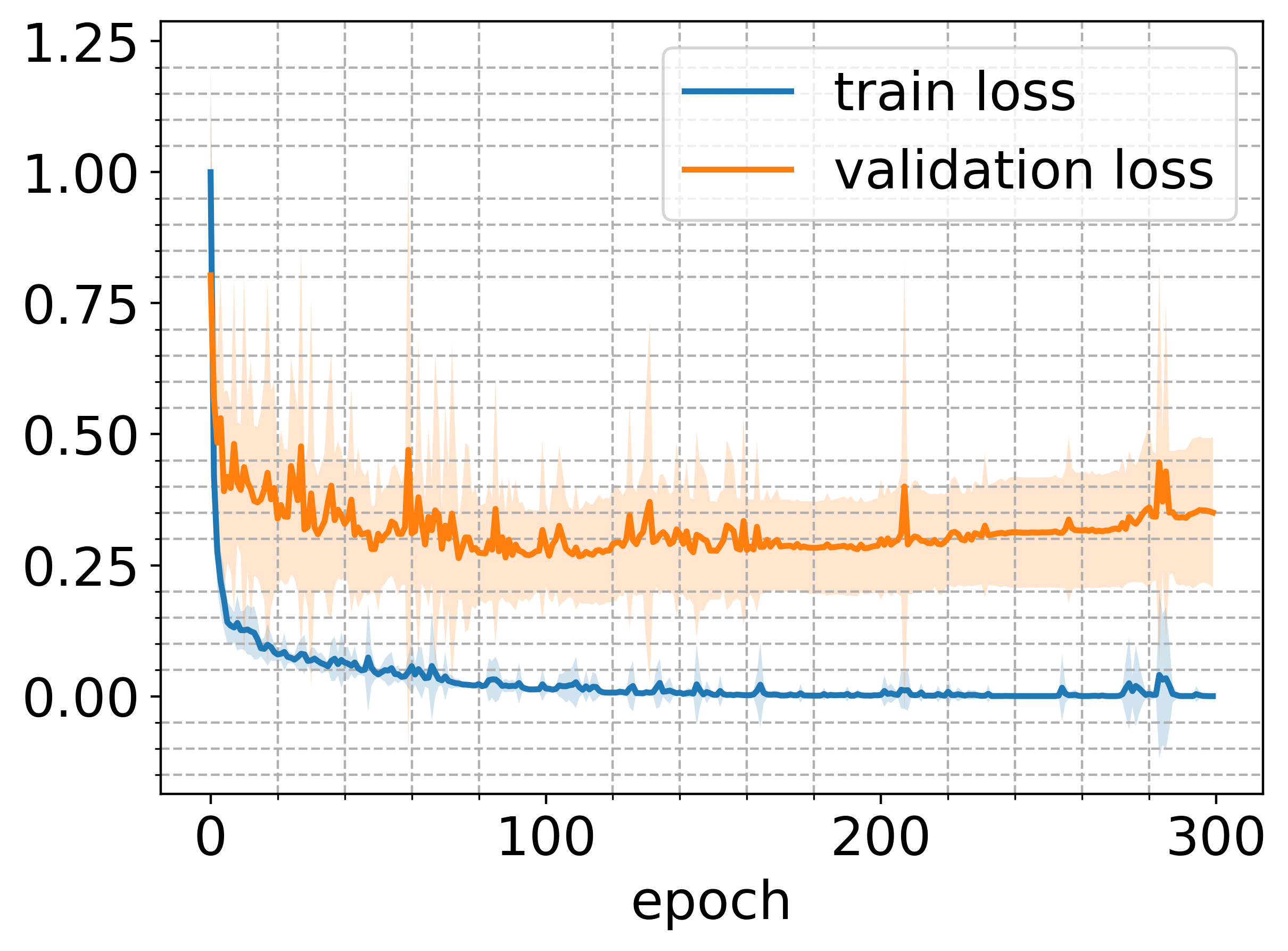}\\
	\centerline{(a)}\medskip\\
	\includegraphics[width=0.48\linewidth]{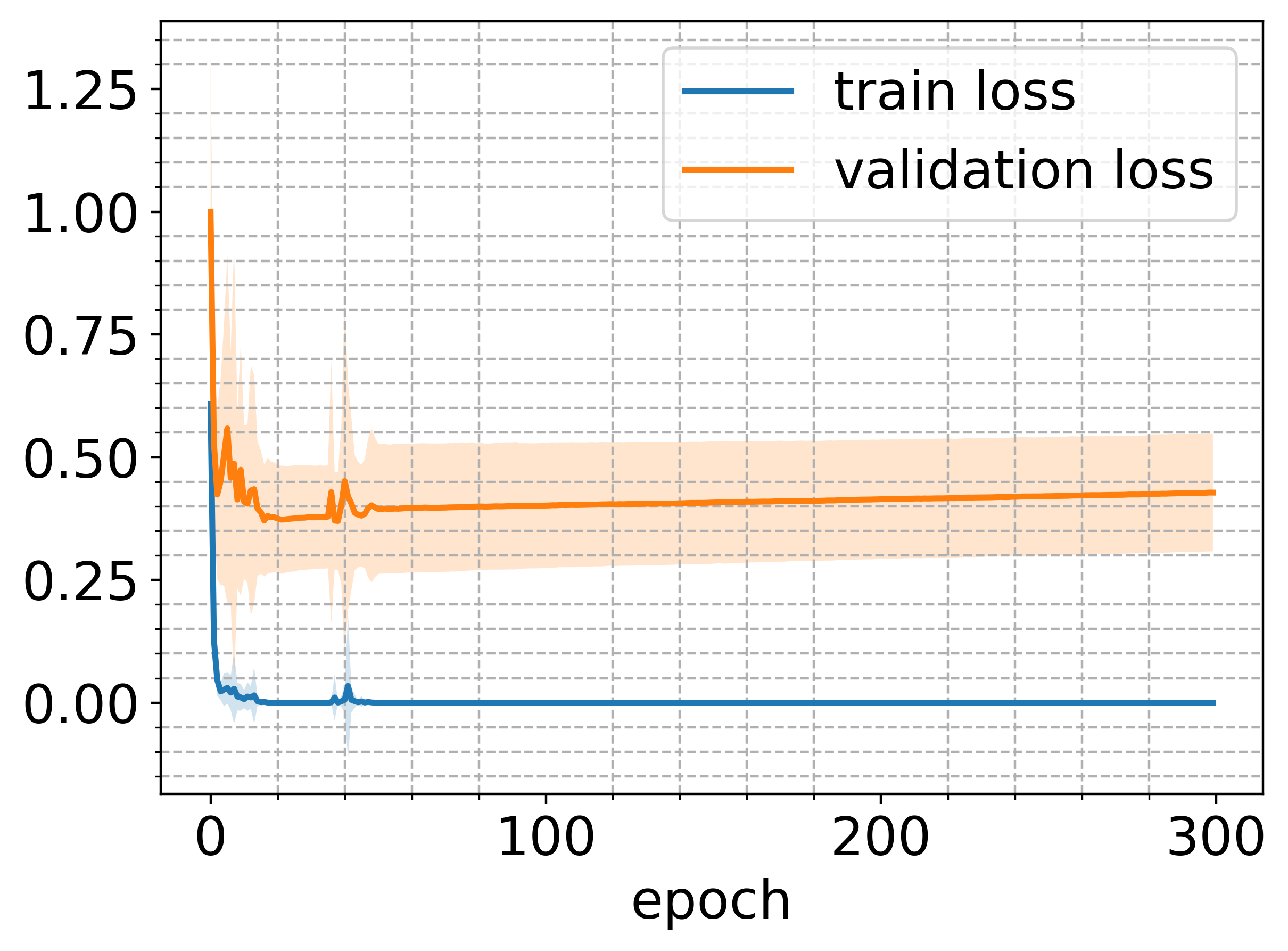}
	\includegraphics[width=0.48\linewidth]{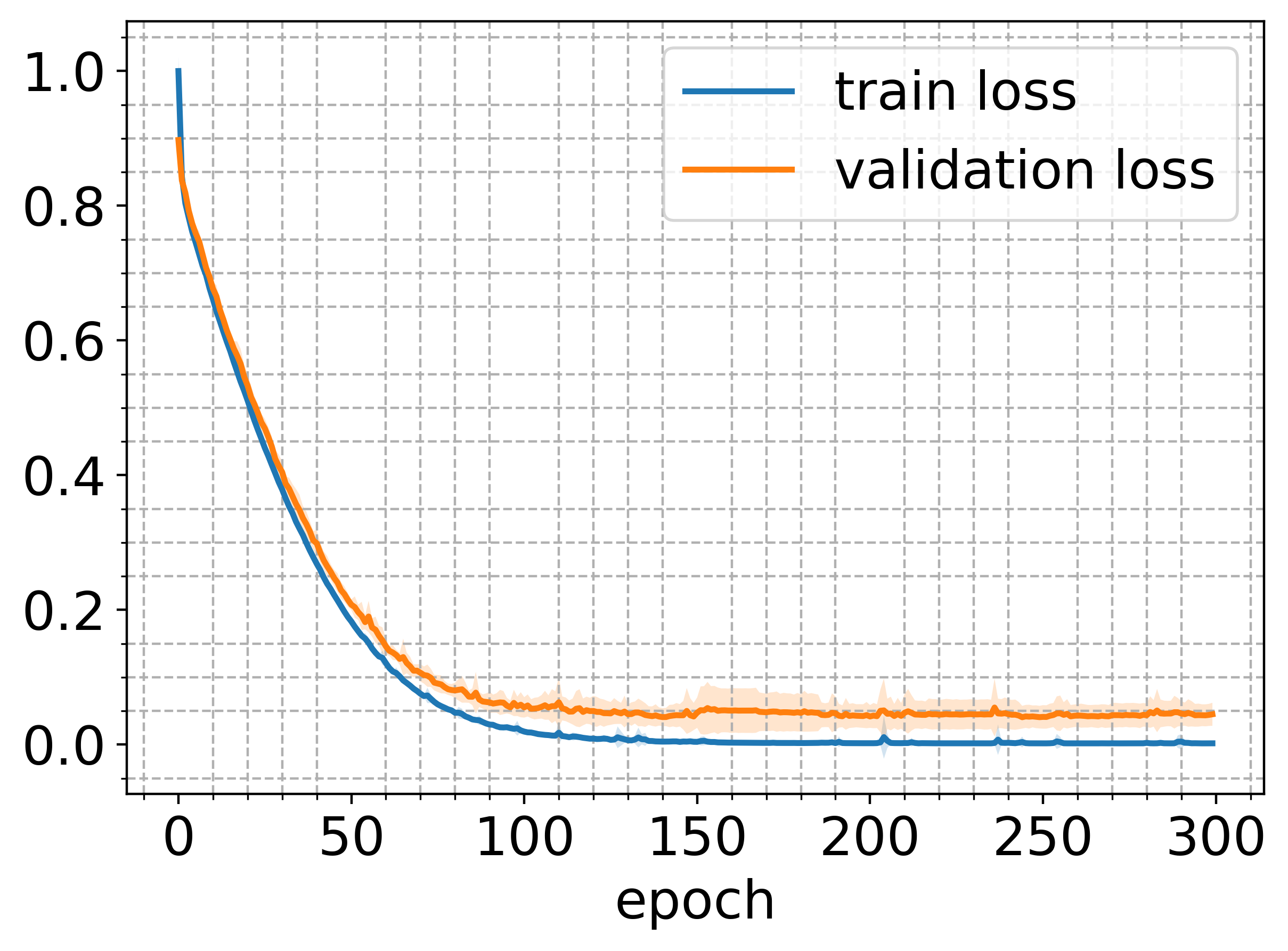}\\
	\centerline{(b)}\medskip\\
	\includegraphics[width=0.48\linewidth]{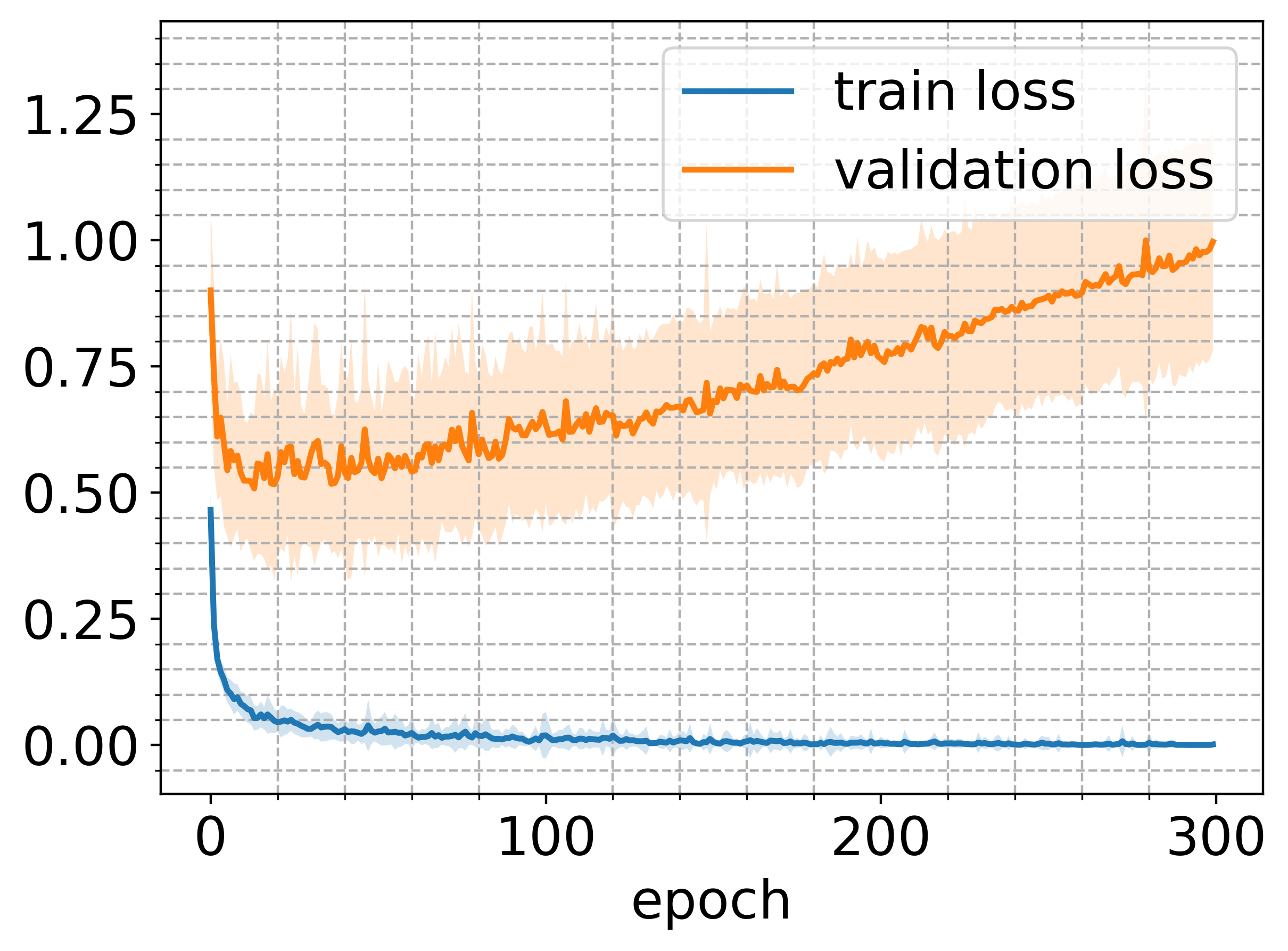}
	\includegraphics[width=0.48\linewidth]{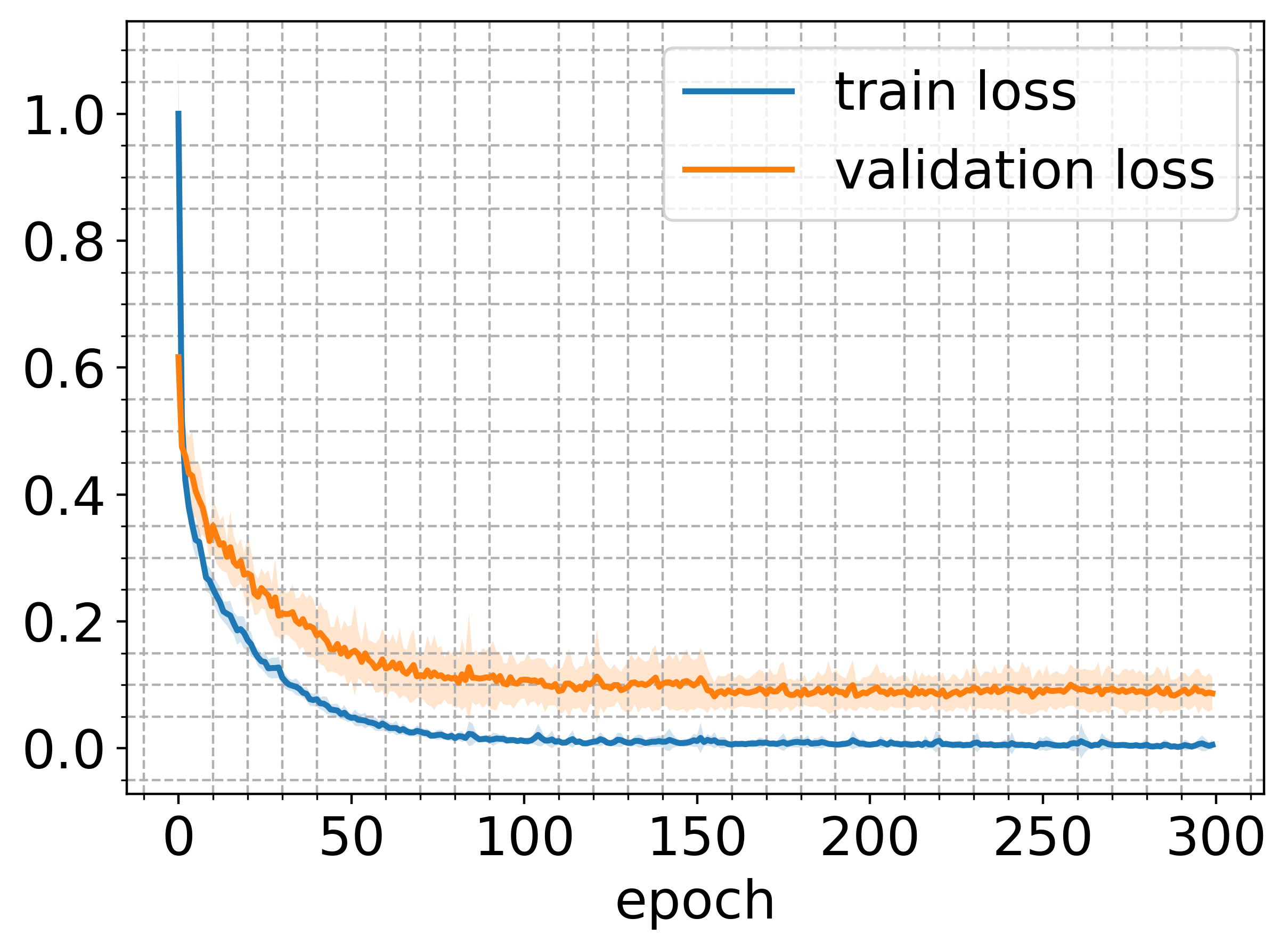}\\
	\centerline{(c)}\medskip\\
	
	\caption{Train and validation loss of CNN (left) and Bayesian NN (right) on (a) Pavia Centre, (b) Botswana, and (c) Salinas datasets.}
	\label{fig:alllosstrain}
\end{figure}

\subsection{Experiment I: Classification Results} \label{sssec:exp1}
Table~\ref{tab:trainresultsall} reports classification results for all comparison methods. The BNN performs better than the CNN for all datasets and outperforms the Random Forest.
For the Botswana and Pavia datasets, BNN mean Kappa score outperforms the CNN by $0.032$ and $0.025$ units. CNN and BNN perform comparably on the Salinas dataset.
Additionally, the BNN achieves a lower variance of the scores in the Botswana and Pavia datasets, which implies a more stable performance across the test runs.
Performing only a single network draw achieves already a reasonably good performance for the BNN.
However, ensemble learning with $50$ sample networks further improves the performance.
A detailed investigation of the class-based models' prediction results are provided in the supplementary materials, in Tab. \ref{tab:class_acc_botswana}, \ref{tab:class_acc_pavia}, and  \ref{tab:class_acc_salinas}.

Studying the evolution of the train and validation loss during
training may reveal indicators for network overfitting.
Figure~\ref{fig:alllosstrain} shows the losses for the CNN on the left and the
BNN on the right on the Pavia Centre, Botswana, and Salinas datasets. The plot
lines show the means and standard deviations of the loss values over all
experiments.
In all cases, the Bayesian model exhibits a more desirable behavior.
In the Salinas dataset, after around $20$ epochs, the CNN validation loss starts to increase while the training loss keeps decreasing.
This indicates the network cannot generalize anymore and is already overfitting, whereas the Bayesian model validation loss keeps decreasing or remains almost constant, even until epoch $300$.
This indicates that the BNN is less prone to overfitting.
Similar behavior can be observed for the Botswana dataset. The loss curves for the Pavia dataset are quite comparable for the BNN and CNN models.

\begin{figure}
	\begin{minipage}[b]{0.24\linewidth}
		\centering
		\centerline{\includegraphics[width=1\linewidth]{images/labels_Salinas_gt.mat.png}}
		\centerline{(a)}\medskip
	\end{minipage}
	\begin{minipage}[b]{0.24\linewidth}
		\centering
		\centerline{\includegraphics[width=1\linewidth]{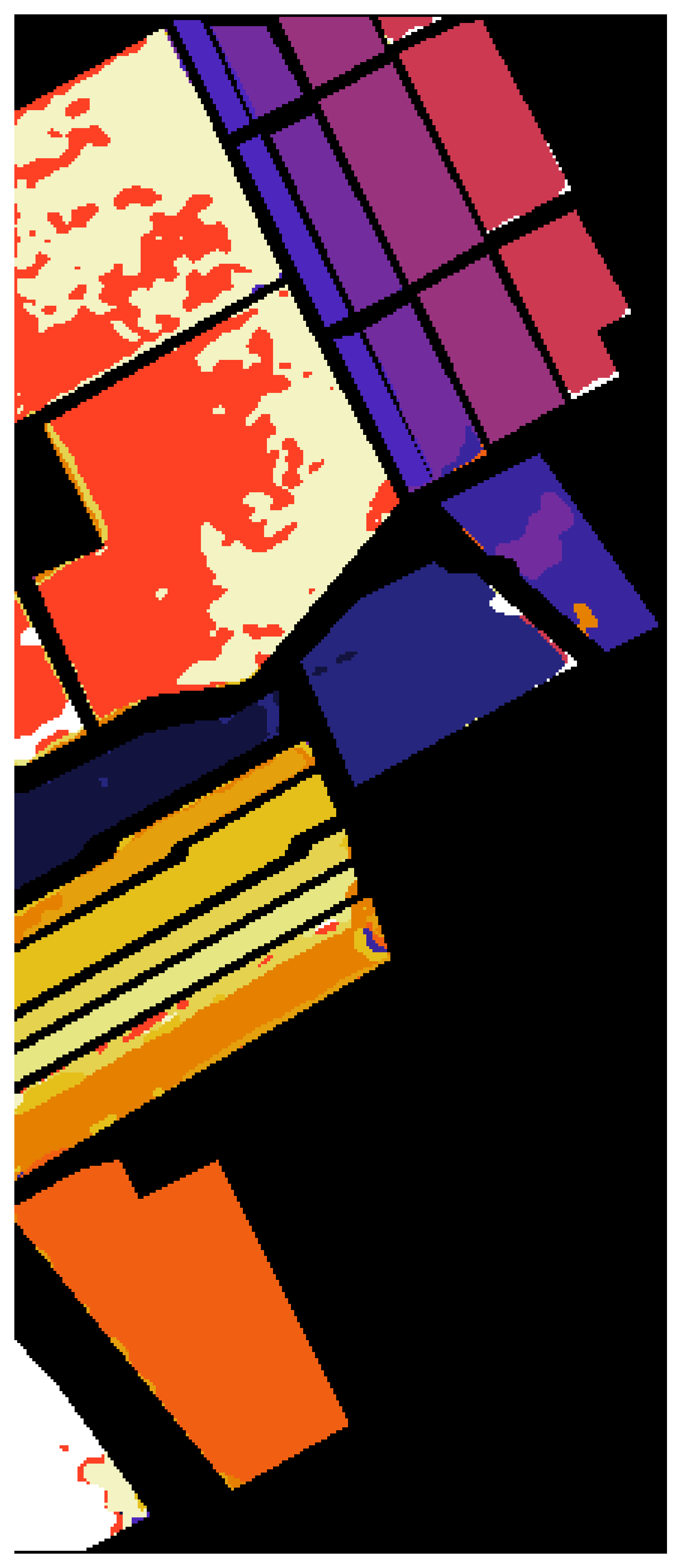}}
		\centerline{(b)}\medskip
	\end{minipage}
	\begin{minipage}[b]{0.24\linewidth}
		\centering
		\centerline{\includegraphics[width=1\linewidth]{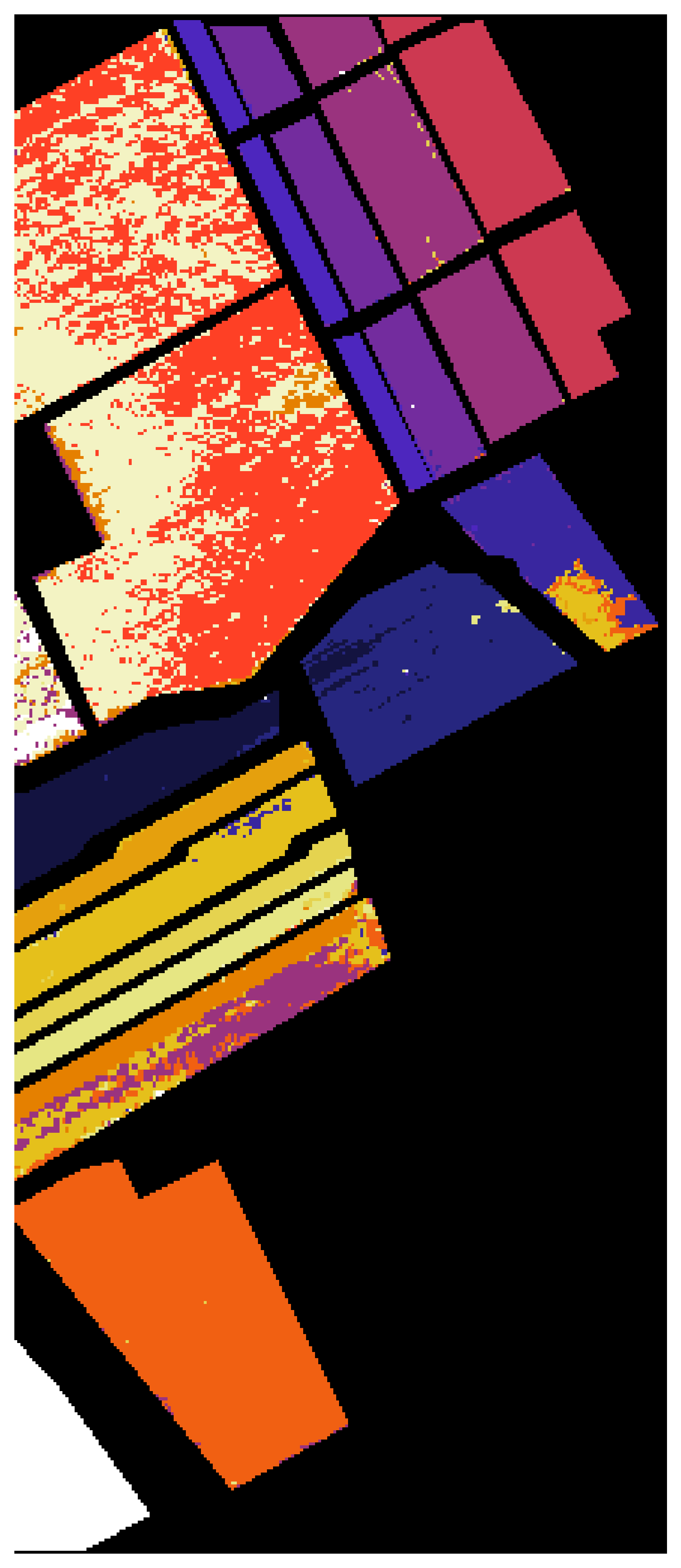}}
		\centerline{(c)}\medskip
	\end{minipage}
	\begin{minipage}[b]{0.24\linewidth}
		\centering
		\centerline{\includegraphics[width=1\linewidth]{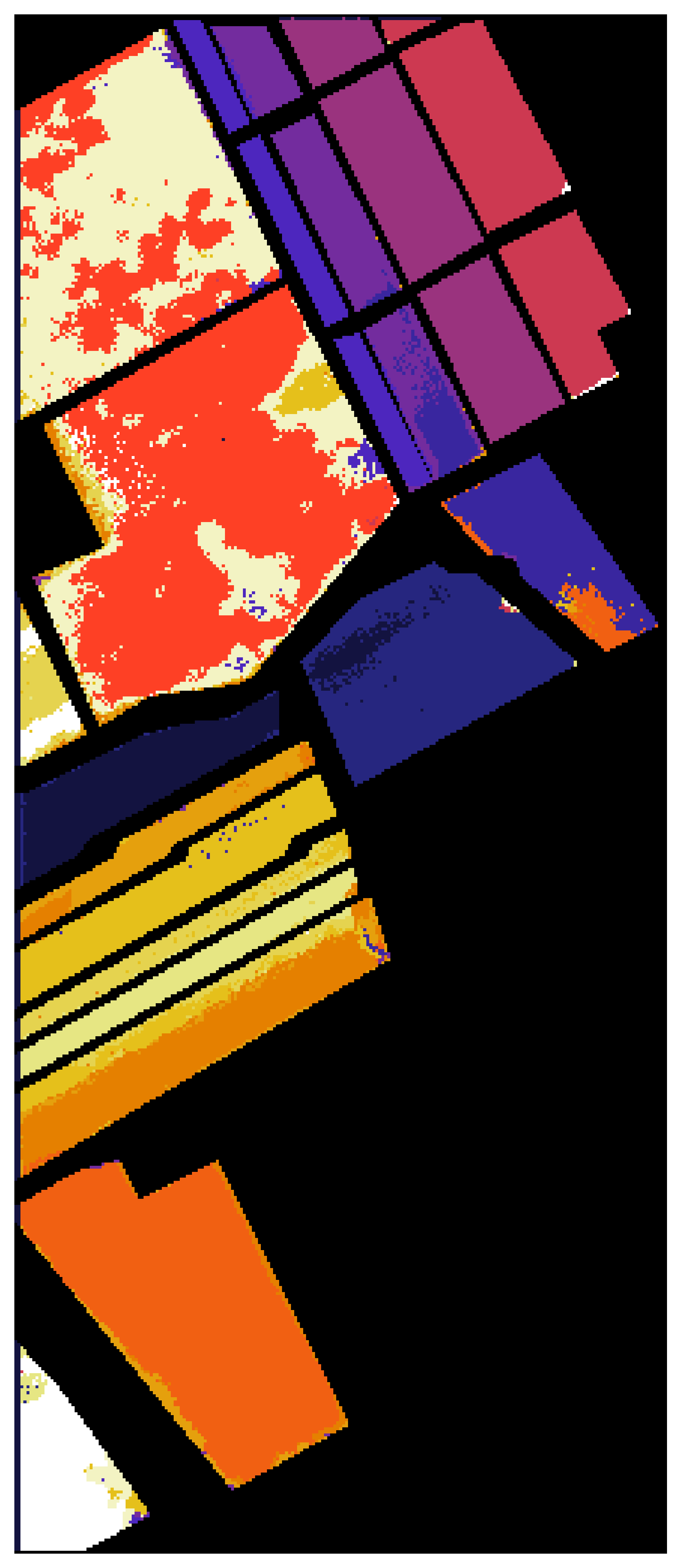}}
		\centerline{(d)}\medskip
	\end{minipage}
	\caption{Prediction of trained Bayesian networks on Salinas dataset. (a) Ground-truth, (b) CNN (c) Random Forest, (d) Bayesian (BNN) predictions.}
	\label{fig:pred_all_Salinas}	
\end{figure}

\begin{figure}
	\begin{minipage}[b]{0.24\linewidth}
		\centering
		\centerline{\includegraphics[width=1\linewidth]{images/labels_Pavia_gt.mat.png}}
		\centerline{(a)}\medskip
	\end{minipage}
	\begin{minipage}[b]{0.24\linewidth}
		\centering
		\centerline{\includegraphics[width=1\linewidth]{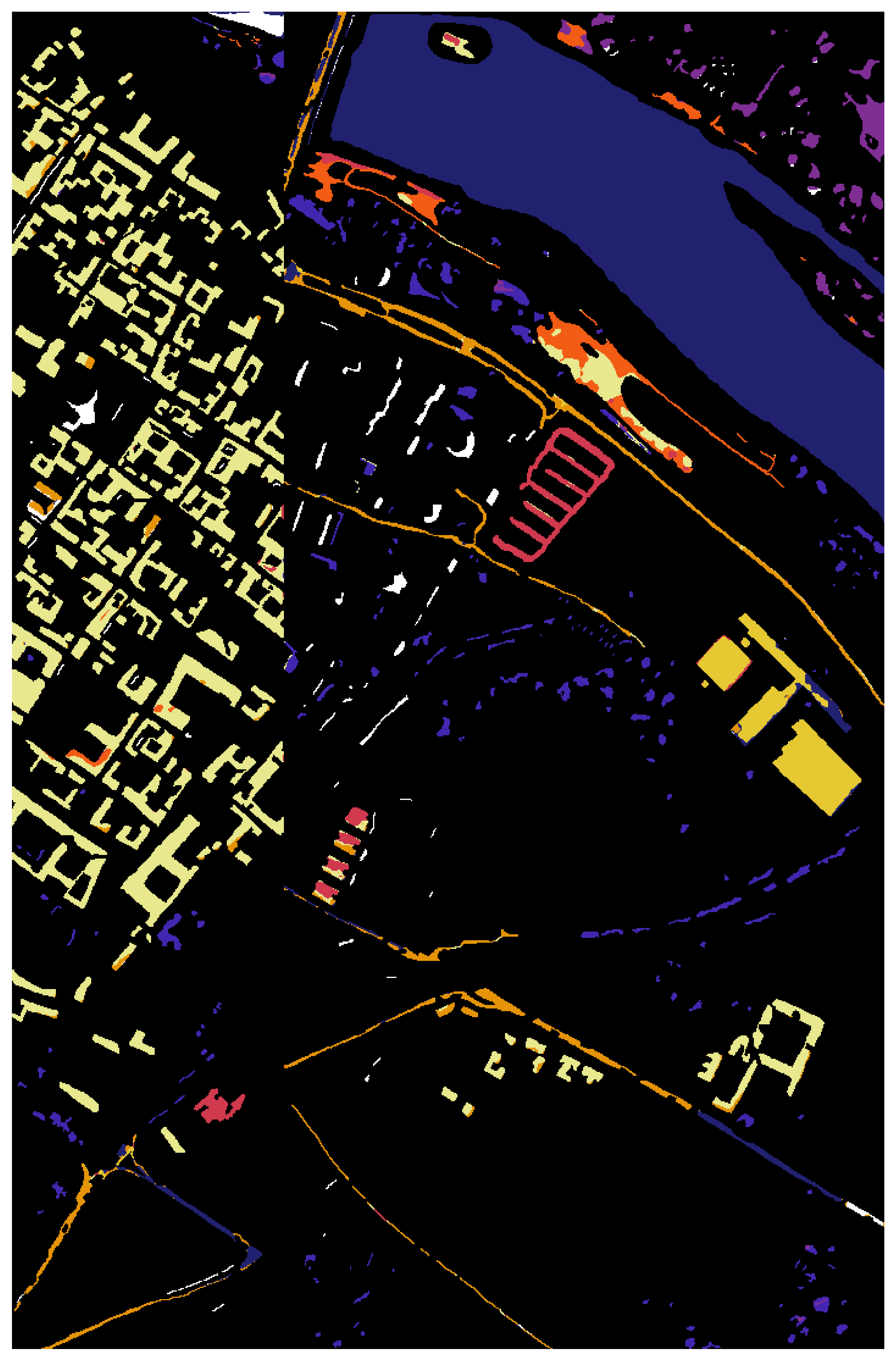}}
		\centerline{(b)}\medskip
	\end{minipage}
	\begin{minipage}[b]{0.24\linewidth}
		\centering
		\centerline{\includegraphics[width=1\linewidth]{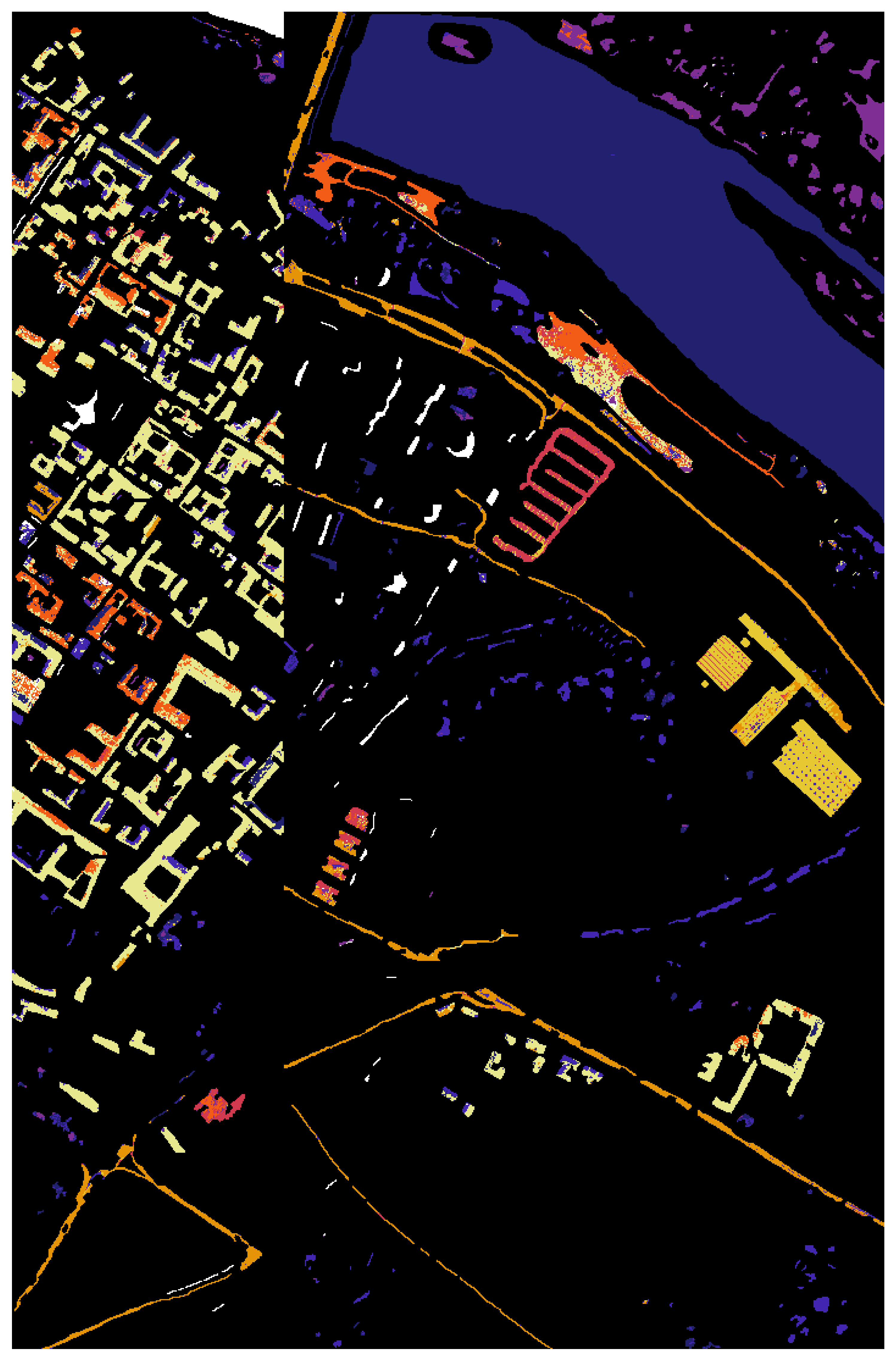}}
		\centerline{(c)}\medskip
	\end{minipage}
	\begin{minipage}[b]{0.24\linewidth}
		\centering
		\centerline{\includegraphics[width=1\linewidth]{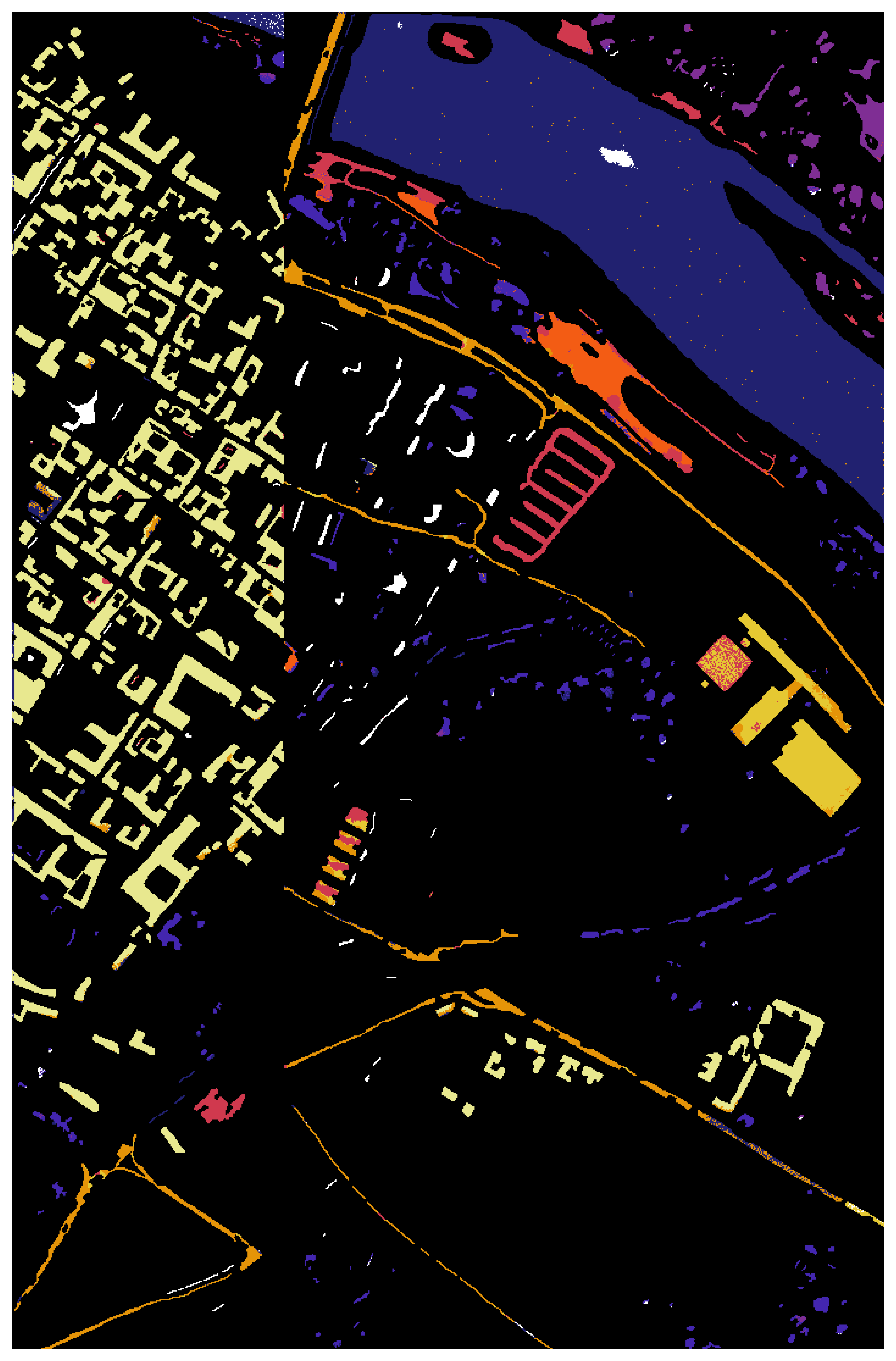}}
		\centerline{(d)}\medskip
	\end{minipage}
	\caption{Prediction of trained Bayesian networks on Pavia dataset. (a) Ground-truth, (b) CNN (c) Random Forest, (d) Bayesian (BNN) predictions.}
	\label{fig:pred_all_Pavia}	
\end{figure}

Qualitative results on the Pavia and Salinas datasets are shown in Fig.~\ref{fig:pred_all_Salinas} and Fig.~\ref{fig:pred_all_Pavia}.
Analogously to the results in Tab.~\ref{tab:trainresultsall}, the Bayesian architecture outperforms the CNN model.
This can be particularly well be observed in the large homogeneous areas of the Salinas dataset.

\subsection{Experiment II: Robustness against Pruning}
\begin{figure}
	\begin{minipage}[b]{.49\linewidth}
		\centering
		\centerline{\includegraphics[width=1\linewidth]{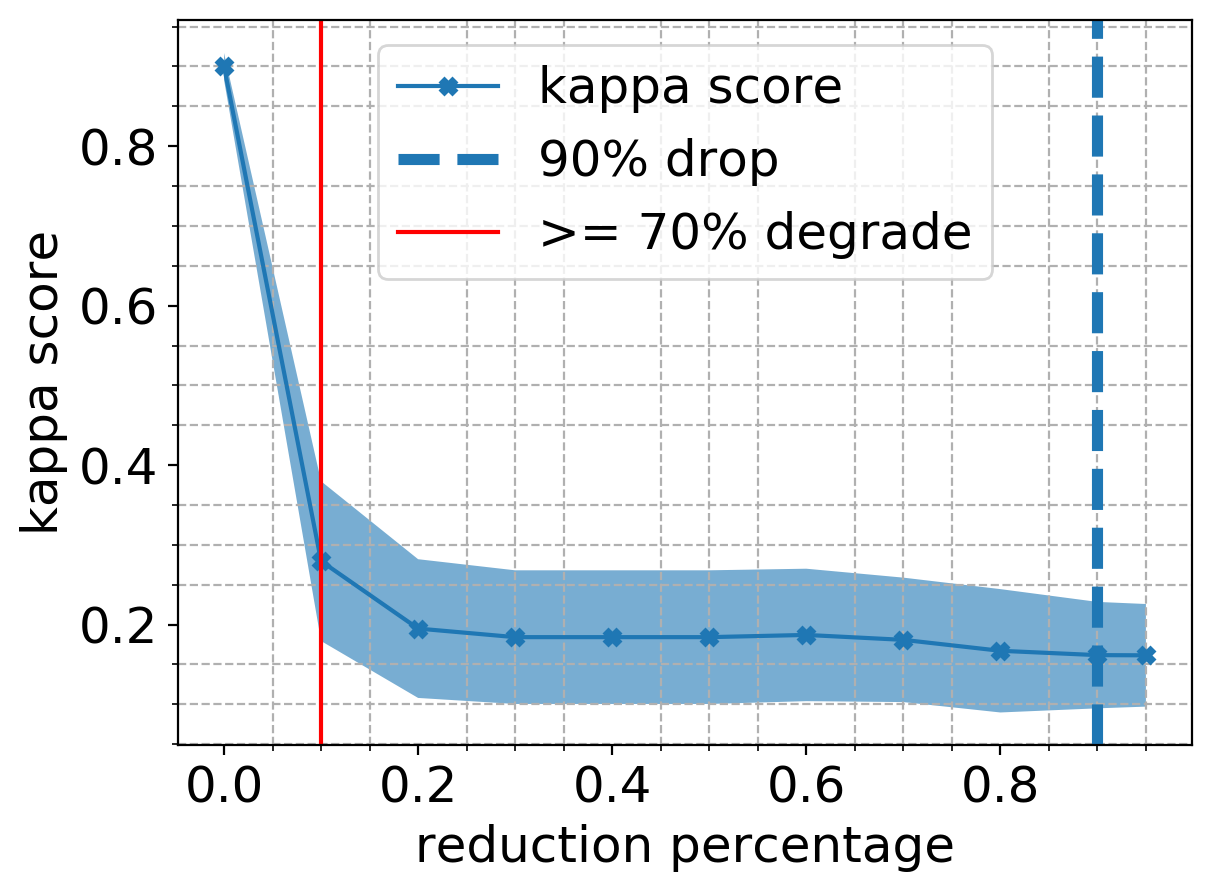}}
		\centerline{(a)}\medskip
	\end{minipage}
	\begin{minipage}[b]{.49\linewidth}
		\centering
		\centerline{\includegraphics[width=1\linewidth]{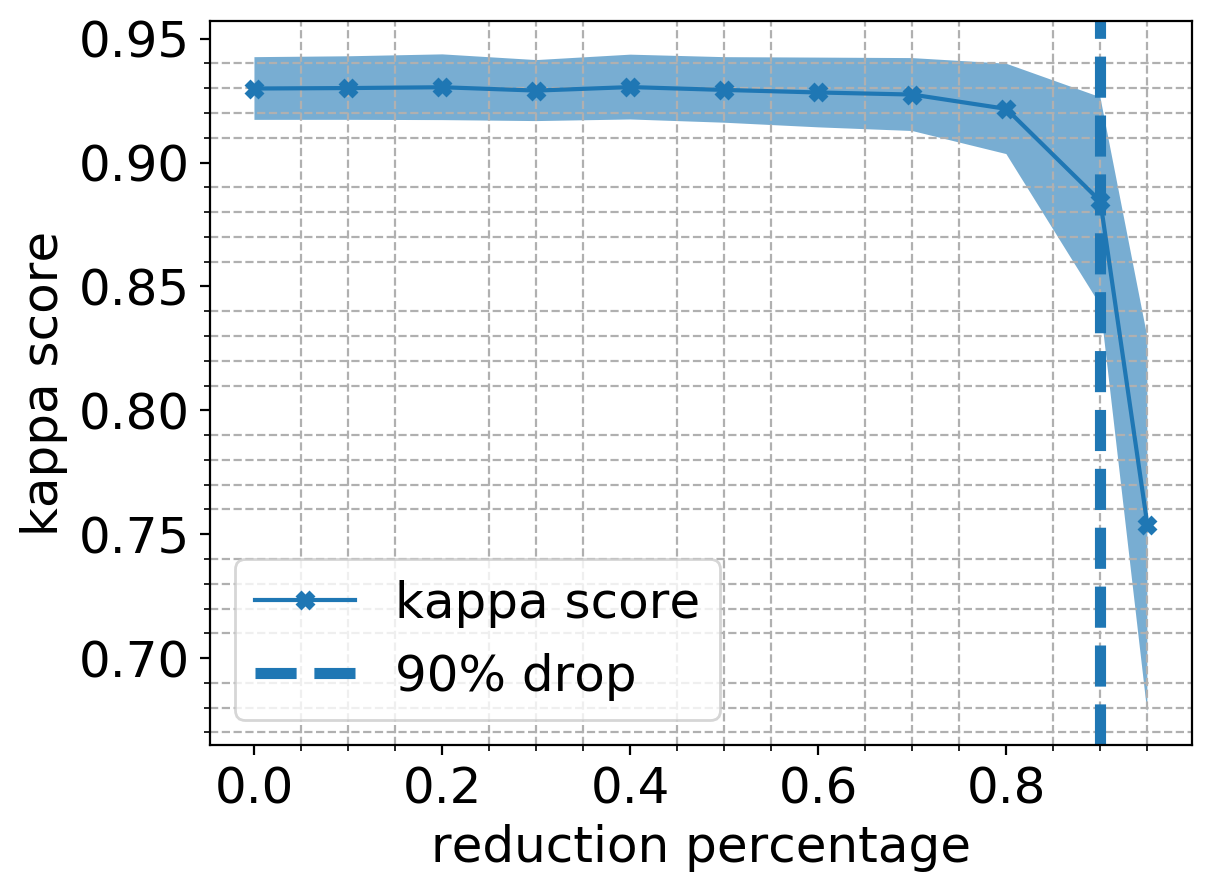}}
		\centerline{(b)}\medskip
	\end{minipage}
	\begin{minipage}[b]{.49\linewidth}
		\centering
		\centerline{\includegraphics[width=1\linewidth]{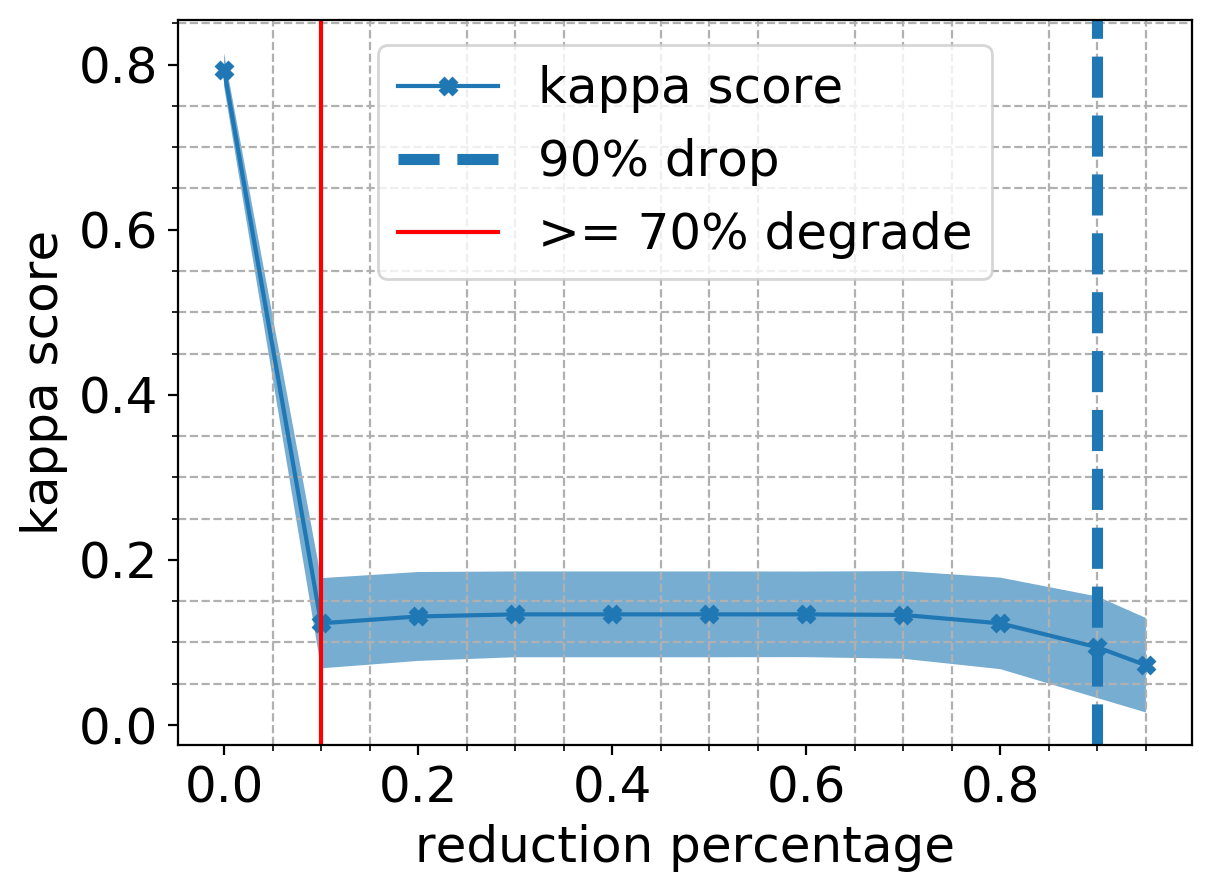}}
		\centerline{(c)}\medskip
	\end{minipage}
	\begin{minipage}[b]{.49\linewidth}
		\centering
		\centerline{\includegraphics[width=1\linewidth]{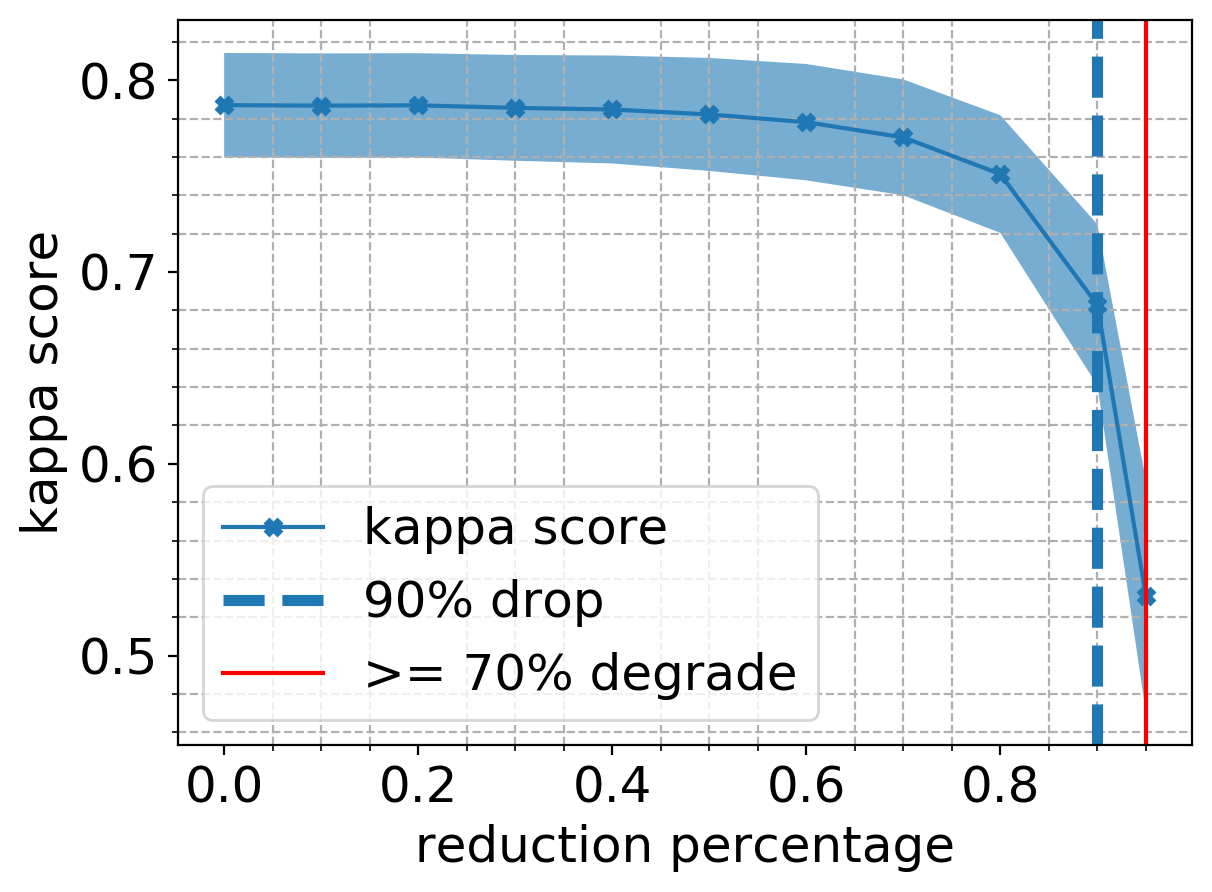}}
		\centerline{(d)}\medskip
	\end{minipage}
	\begin{minipage}[b]{.49\linewidth}
		\centering
		\centerline{\includegraphics[width=1\linewidth]{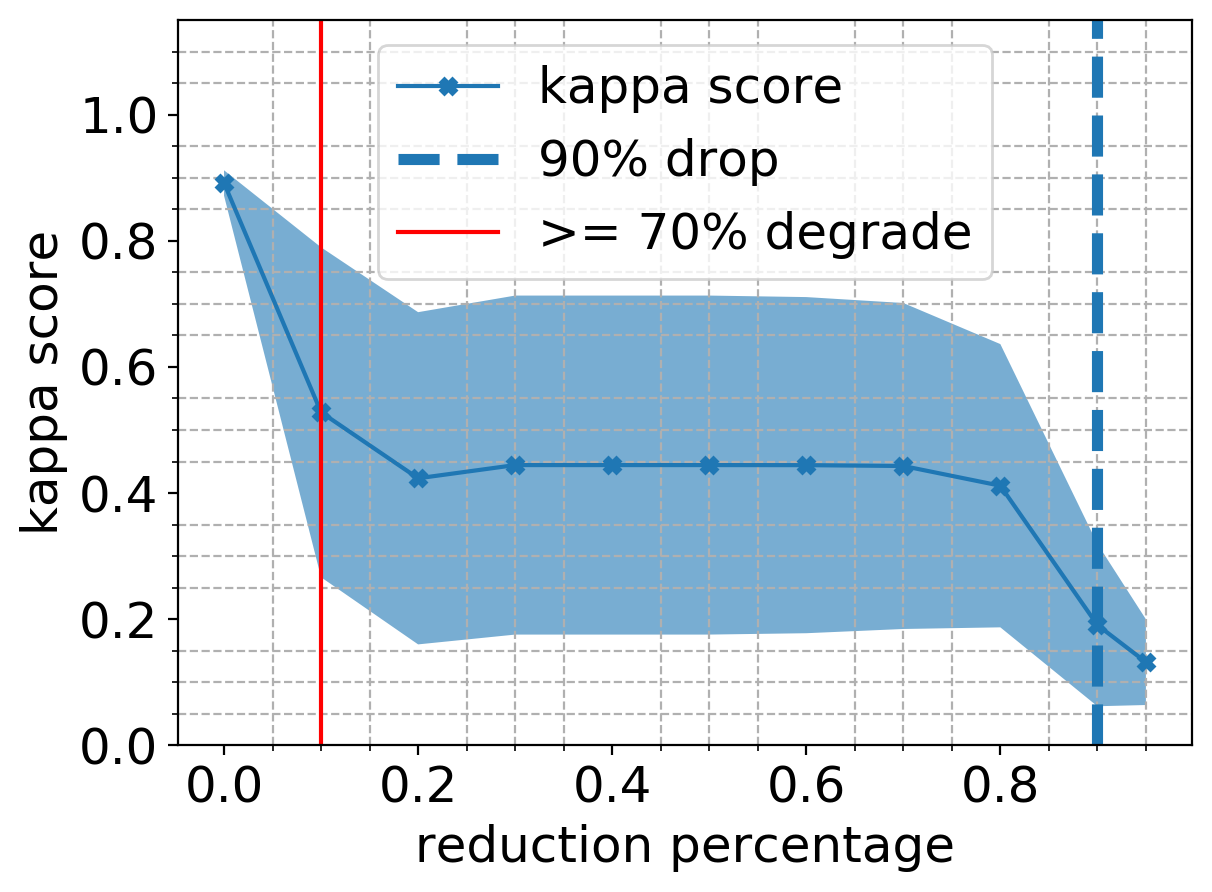}}
		\centerline{(e)}\medskip
	\end{minipage}
	\begin{minipage}[b]{.49\linewidth}
		\centering
		\centerline{\includegraphics[width=1\linewidth]{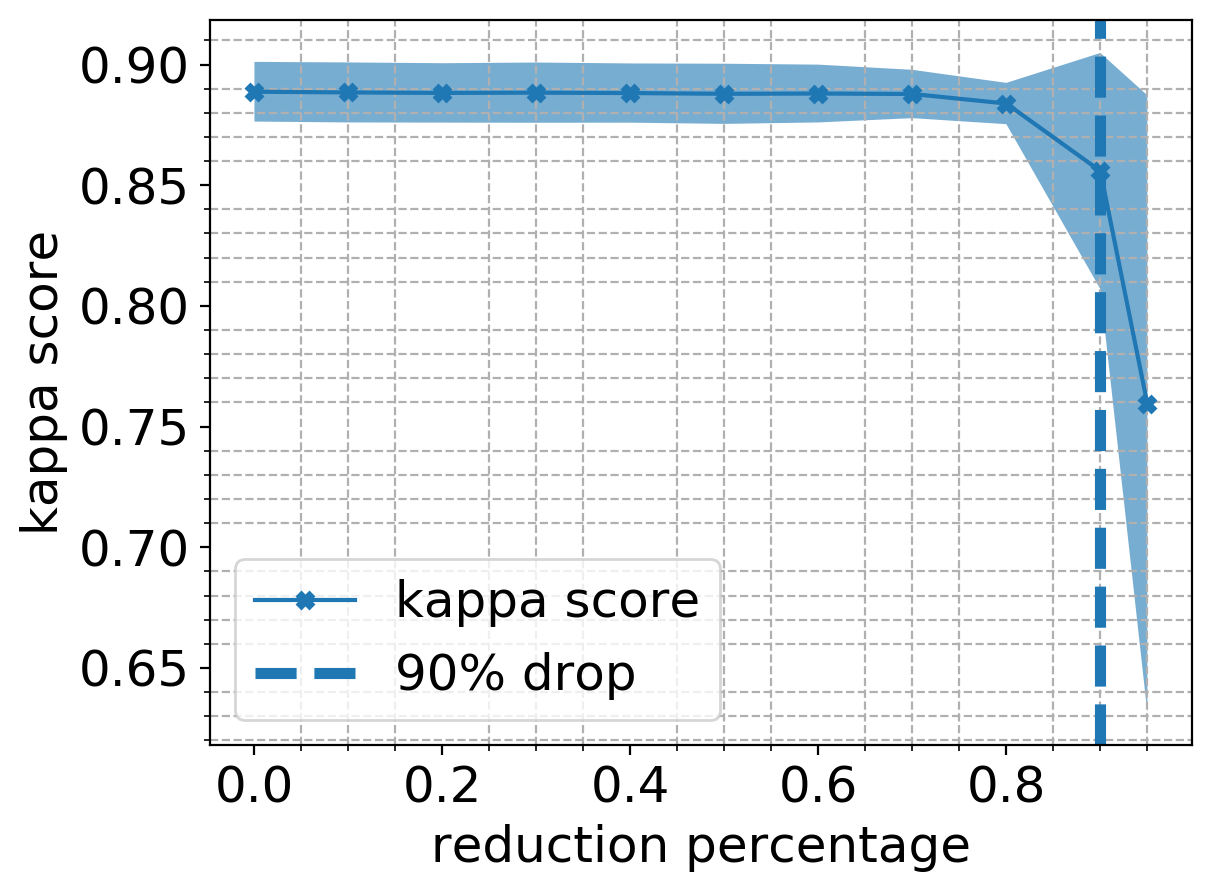}}
		\centerline{(f)}\medskip
	\end{minipage}
	\caption{Kappa score after global pruning of the networks. The left column shows the CNN and the right column shows the BNN. (a) and (b) are computed on Botswana, (c) and (d) on Salinas, and (e) and (f) on the Pavia dataset. }
	\label{fig:prune_all}
\end{figure}

This pruning experiment is performed on the models that were trained for Experiment I. 
We iteratively prune $10\%$ of the available network weights. More in detail, CNN pruning is performed by pruning the weights with the smallest magnitude (absolute value). BNN pruning is performed by pruning the weights with the highest probability of being zero.
This process is done for all the experiments, and the mean and the standard deviation of the kappa scores are provided in Fig.~\ref{fig:prune_all}.
In this figure, the red line indicates where the model reaches below $70\%$ of its initial kappa score, and the blue dashed line indicates where $90\%$ of the weights have been filtered. 

\begin{figure*}
	\begin{minipage}[b]{.32\linewidth}
		\centering
		\centerline{\includegraphics[width=1\linewidth]{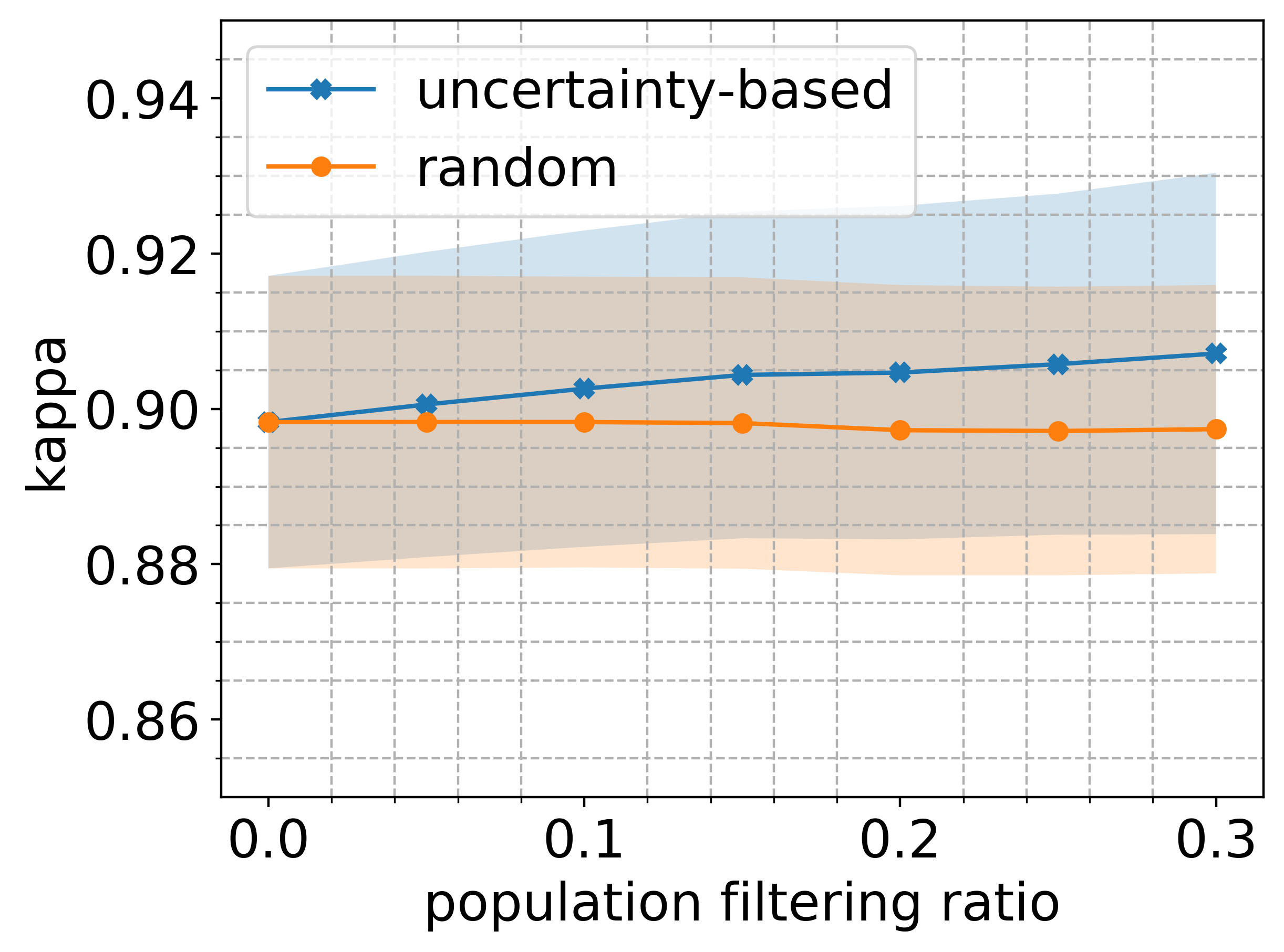}}
		\centerline{(a)}\medskip
	\end{minipage}
	\begin{minipage}[b]{.32\linewidth}
		\centering
		\centerline{\includegraphics[width=1\linewidth]{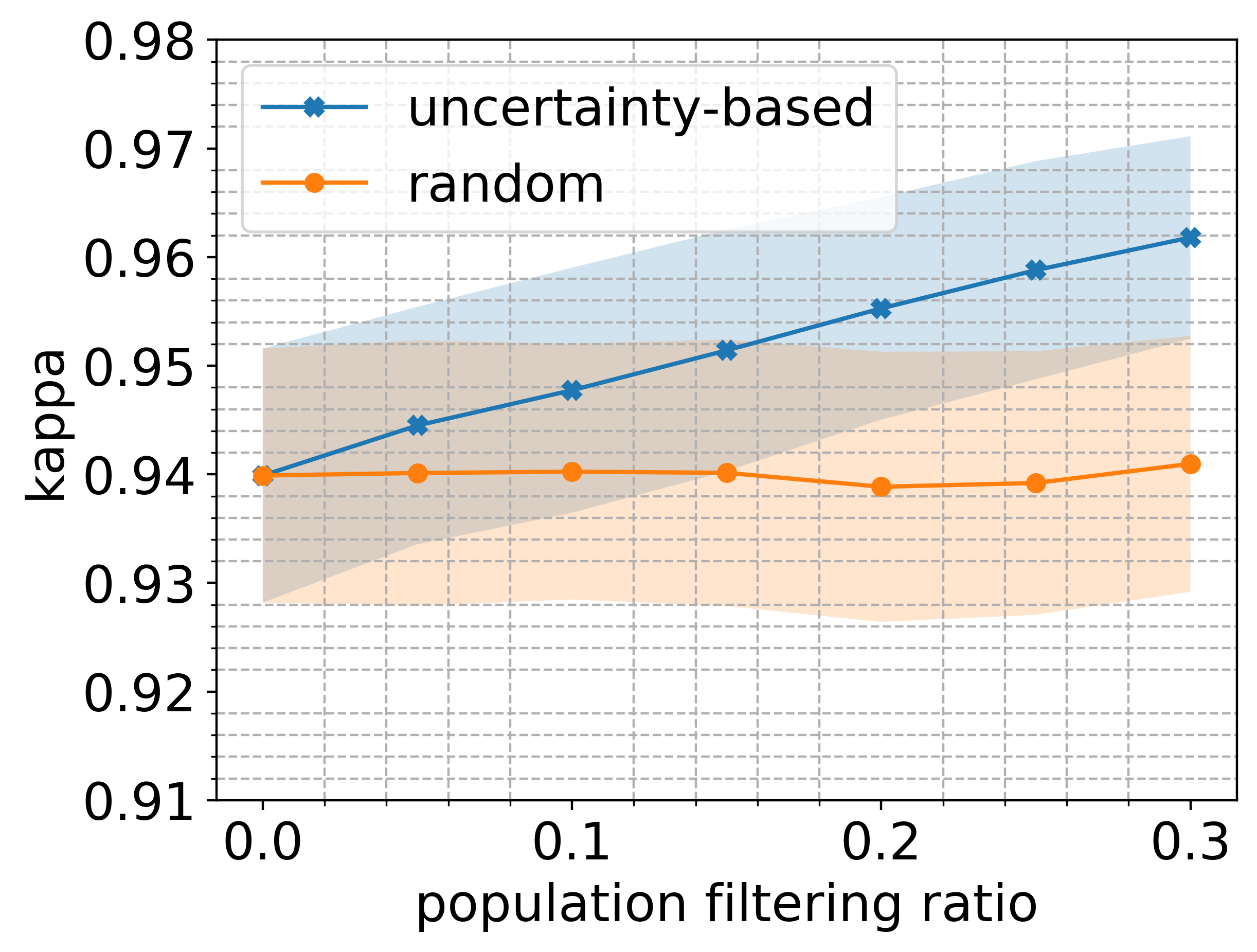}}
		\centerline{(b)}\medskip
	\end{minipage}
	\begin{minipage}[b]{.32\linewidth}
		\centering
		\centerline{\includegraphics[width=1\linewidth]{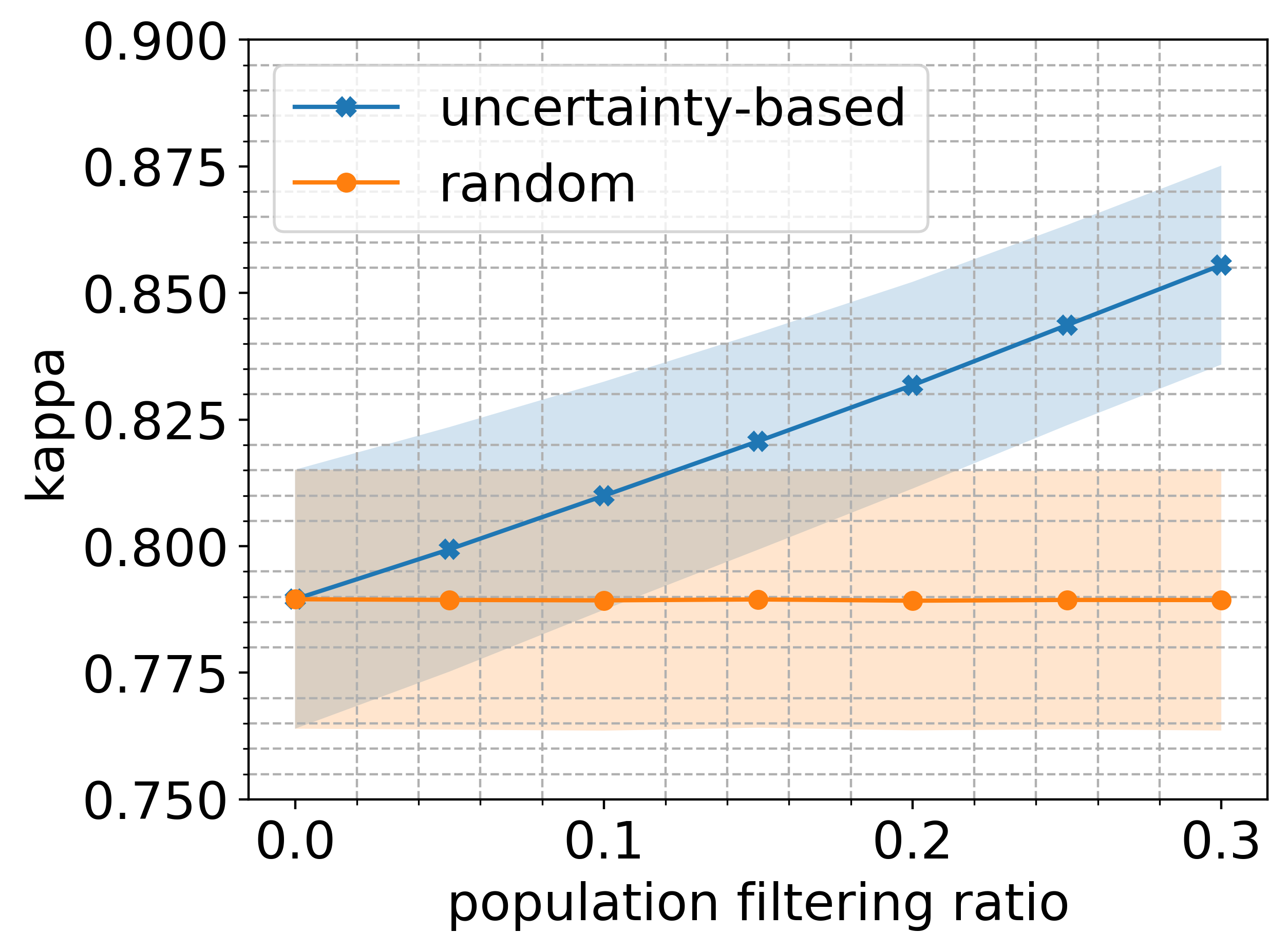}}
		\centerline{(c)}\medskip
	\end{minipage}
	
	\caption{Kappa score versus uncertain data filtering on (a) Pavia, (b) Botswana, and (c) Salinas datasets. Filtering out the uncertain test samples improves the classification performance, which indicates that the uncertainty has a positive correlation with the prediction error.}
	\label{fig:uncert_vs_kappa}
\end{figure*}

\begin{figure}
	\begin{minipage}[b]{.32\linewidth}
		\centering
		\centerline{\includegraphics[width=1\linewidth]{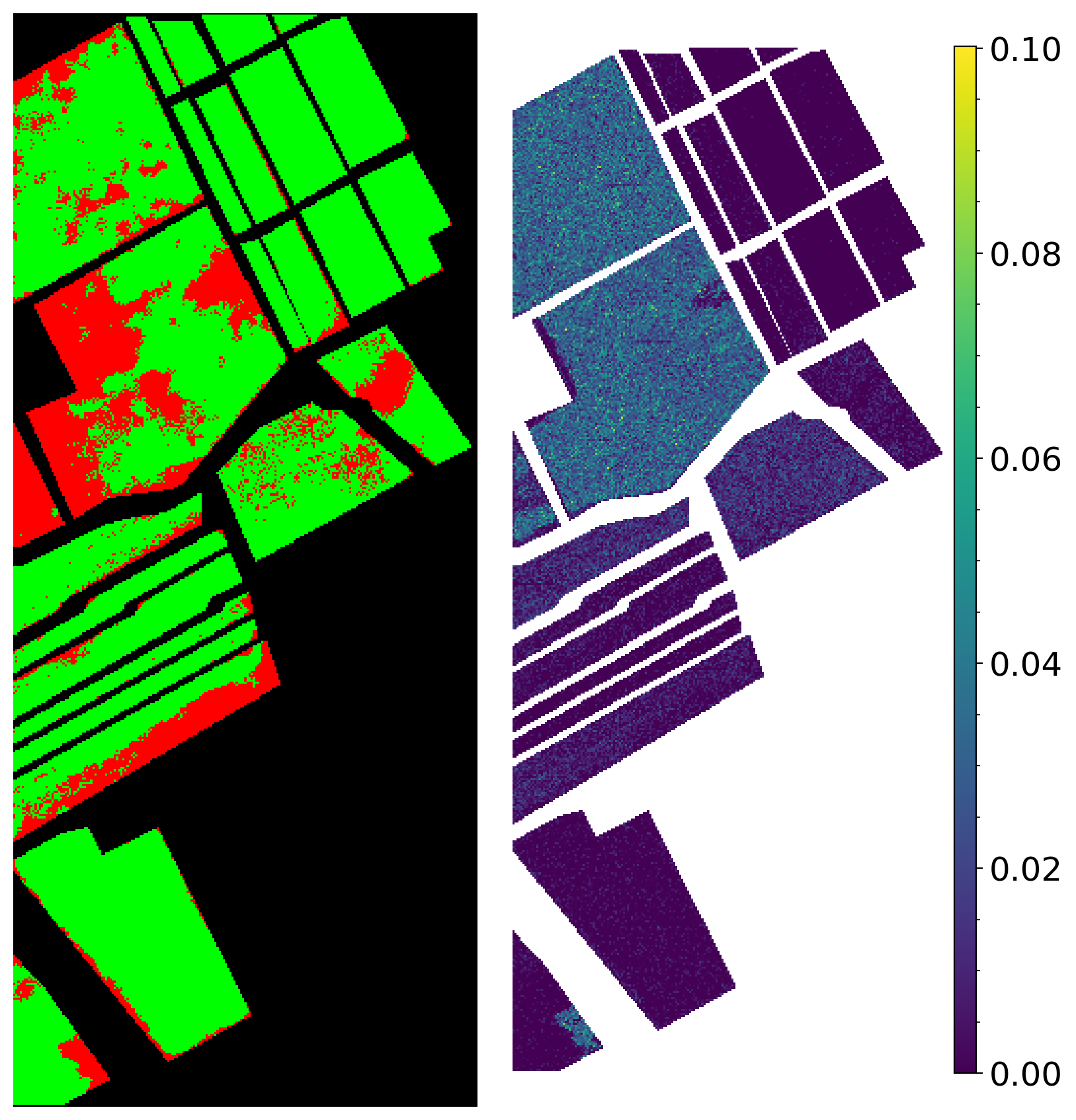}}
		\centerline{(a)}\medskip
	\end{minipage}
	\begin{minipage}[b]{.32\linewidth}
		\centering
		\centerline{\includegraphics[width=1\linewidth]{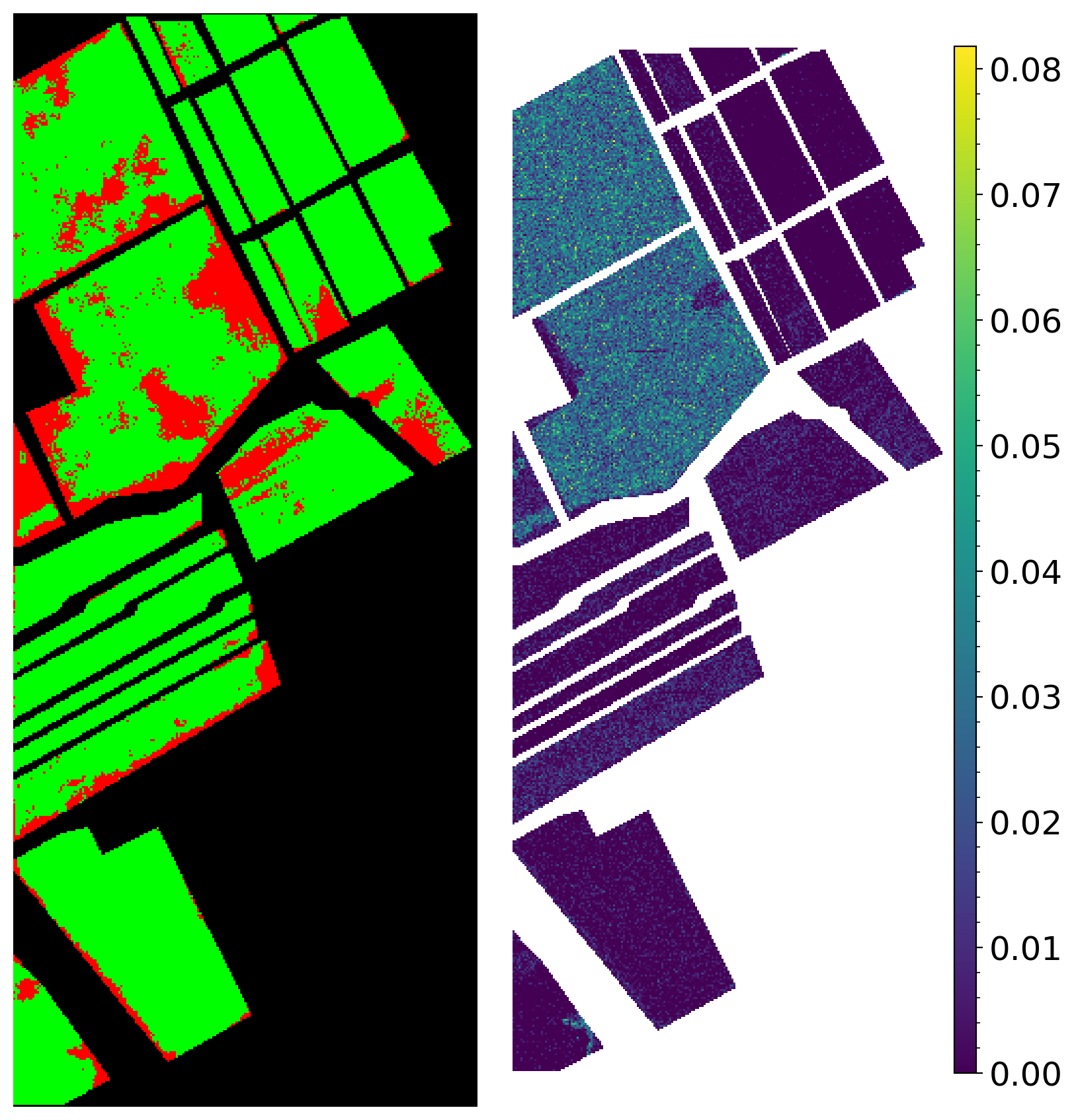}}
		\centerline{(b)}\medskip
	\end{minipage}
	\begin{minipage}[b]{.32\linewidth}
		\centering
		\centerline{\includegraphics[width=1\linewidth]{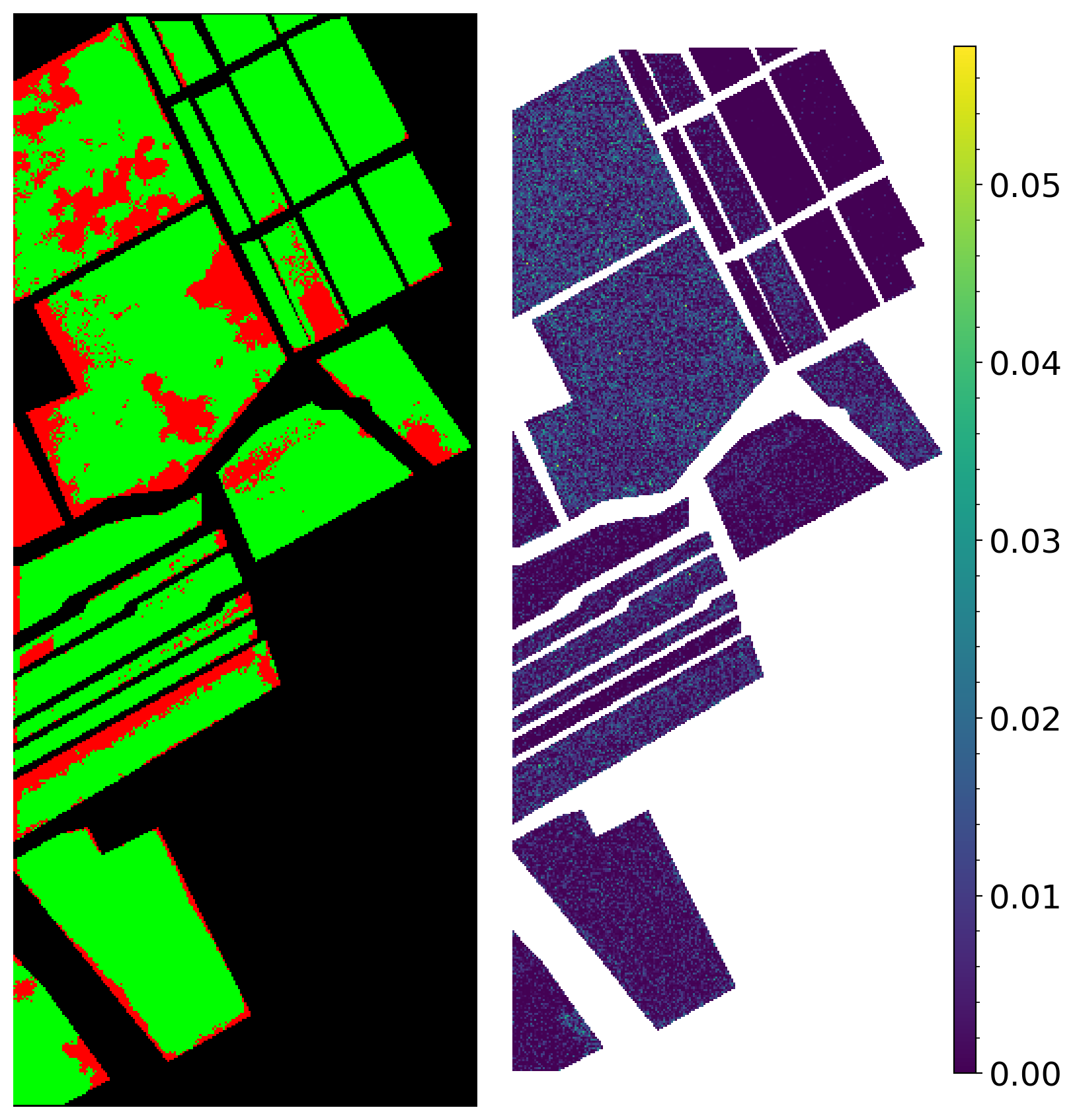}}
		\centerline{(c)}\medskip
	\end{minipage}

	\caption{Prediction correctness (left) versus the uncertainty (right) for three Bayesian networks trained on Salinas dataset.}
	\label{fig:cor_uncert_Salinas}
\end{figure}

The results are similar across all three datasets. The CNN performance deteriorates relatively quickly. The $70\%$ performance degradation is reached when pruning only $10\%$ of the weights.
On the other hand, the kappa score of the BNNs remains high when even large numbers of parameters are pruned. Even after pruning $90\%$ of the network weights, the kappa score is still above $70\%$.
Hence, we conclude that the Bayesian network is more stable, and it can be reduced to a much smaller sub-network. Hence, the Bayesian CNN can maintain the high performance under higher compression ratios and has a higher compression capability.

\subsection{Experiment III: Evaluation of Uncertainties}\label{sec:chap5-ExpIII}
One of the advantages of Bayesian networks over the non-Bayesian networks is the additional uncertainty metric. As discussed in section \ref{sec:background}, there are two types of uncertainties: one accounts for uncertainty in the model prediction, and the other one accounts for the uncertainty caused by the data. In this section, since we want to observe the uncertainty in the input data, we will use the \emph{Aleatoric} uncertainty.

We design the following experiment to examine whether the uncertainty provided by the network is informative and correlates with the data quality.
The quality of the uncertainty metric is assessed by removing a part of the test data.
In one variant, we remove just a random portion of the test data. In a second
variant, we remove the most uncertain test data points. These dataset reduction
approaches are put in relation to the prediction accuracy for the Pavia Centre
and Salinas datasets to visually observe the effectiveness of the uncertainty
metric. 

Figure~\ref{fig:uncert_vs_kappa} shows the results. For all datasets, filtering up to $30\%$ of the uncertain test data points results in a significant improvement in the kappa scores. This indicates that the BNN uncertainty can serve as a predictor for samples that are likely misclassified.

Figure~\ref{fig:cor_uncert_Salinas} 
qualitatively shows on the Salinas dataset how the uncertainty metric can indicate the prediction reliability.
Three trained Bayesian networks are used to visualize the uncertainty. The
outputs of these networks are shown in the subfigures (a), (b), and (c). Each
subfigure consists of a left part and a right part. The left part shows in
green correctly labeled pixels, and in red wrongly labeled pixels. The right
part indicates the associated uncertainty. It can be observed that the areas of
wrong prediction exhibit by tendency also higher uncertainty. 

\section{Conclusion}\label{conclusion}
This paper investigated the Bayesian convolutional neural network for HSRS image classification with severely limited training data and compared it with a similar frequentist CNN and an off-the-shelf Random Forest. Bayesian methods, in theory, are more robust against overfitting, especially in the case of limited data. The experimental results in this work demonstrated that a Bayesian network outperforms a similar frequentist CNN and Random Forest. Furthermore, the loss plots illustrate that the Bayesian CNN is, for all datasets, more robust against overfitting compared to the non-Bayesian CNN. Moreover, through a set of pruning and re-validation experiments, the Bayesian network has been shown to converge to a considerably smaller and more compact sub-network. This makes them more suitable to be used on devices with less memory, e.g., embedded chips and FPGAs.

In another aspect, the uncertainty provided by the Bayesian network has been studied. It has been observed that the \emph{Aleatoric} uncertainty has a direct correlation with the prediction error: Filtering out uncertain test data enhances the model's performance on the remainder of the test set, whereas randomly filtering the test data does not have any impact on the model's performance. It indicates that the uncertainty metric correlates with the prediction error, and depending on the application, it can be used to enhance the model's prediction.

\bibliographystyle{ieeetr}
\bibliography{References}

\begin{thebibliography}{10}

\bibitem{hsrs_app1}
S.~Valero, P.~Salembier, and J.~Chanussot, ``Hyperspectral image representation
  and processing with binary partition trees,'' {\em IEEE Transactions on Image
  Processing}, vol.~22, no.~4, pp.~1430--1443, 2013.

\bibitem{hsrs_app2}
C.-I. Chang, {\em Hyperspectral Imaging: Techniques for Spectral Detection and
  Classification}.
\newblock Plenum Publishing Co., 2003.

\bibitem{hughes}
G.~Hughes, ``On the mean accuracy of statistical pattern recognizers,'' {\em
  IEEE Transactions on Information Theory}, vol.~14, no.~1, pp.~55--63, 1968.

\bibitem{fast_and_efficient}
A.~{Davari}, H.~C. {Özkan}, A.~{Maier}, and C.~{Riess}, ``Fast and efficient
  limited data hyperspectral remote sensing image classification via gmm-based
  synthetic samples,'' {\em IEEE Journal of Selected Topics in Applied Earth
  Observations and Remote Sensing}, vol.~12, no.~7, pp.~2107--2120, 2019.

\bibitem{davari2018gmm}
A.~Davari, E.~Aptoula, B.~Yanikoglu, A.~Maier, and C.~Riess, ``Gmm-based
  synthetic samples for classification of hyperspectral images with limited
  training data,'' {\em IEEE Geoscience and Remote Sensing Letters}, vol.~15,
  no.~6, pp.~942--946, 2018.

\bibitem{lr_lim10}
M.~Chi, R.~Feng, and L.~Bruzzone, ``Classification of hyperspectral
  remote-sensing data with primal svm for small-sized training dataset
  problem,'' {\em Advances in Space Research}, vol.~41, no.~11, pp.~1793--1799,
  2008.

\bibitem{lr_lim11}
Q.~{Jackson} and D.~A. {Landgrebe}, ``An adaptive classifier design for
  high-dimensional data analysis with a limited training data set,'' {\em IEEE
  Transactions on Geoscience and Remote Sensing}, vol.~39, no.~12,
  pp.~2664--2679, 2001.

\bibitem{lr_lim12}
J.~{Xia}, J.~{Chanussot}, P.~{Du}, and X.~{He}, ``Rotation-based support vector
  machine ensemble in classification of hyperspectral data with limited
  training samples,'' {\em IEEE Transactions on Geoscience and Remote Sensing},
  vol.~54, no.~3, pp.~1519--1531, 2016.

\bibitem{davari2015effect}
A.~A. Davari, E.~Aptoula, and B.~Yanikoglu, ``On the effect of synthetic
  morphological feature vectors on hyperspectral image classification
  performance,'' in {\em 2015 23nd Signal Processing and Communications
  Applications Conference (SIU)}, pp.~653--656, IEEE, 2015.

\bibitem{lr_lim20}
X.~{Kang}, X.~{Xiang}, S.~{Li}, and J.~A. {Benediktsson}, ``Pca-based
  edge-preserving features for hyperspectral image classification,'' {\em IEEE
  Transactions on Geoscience and Remote Sensing}, vol.~55, no.~12,
  pp.~7140--7151, 2017.

\bibitem{dalla2010morphological}
M.~Dalla~Mura, J.~A. Benediktsson, B.~Waske, and L.~Bruzzone, ``Morphological
  attribute profiles for the analysis of very high resolution images,'' {\em
  IEEE Transactions on Geoscience and Remote Sensing}, vol.~48, no.~10,
  pp.~3747--3762, 2010.

\bibitem{gan}
I.~J. Goodfellow, J.~Pouget-Abadie, M.~Mirza, B.~Xu, D.~Warde-Farley, S.~Ozair,
  A.~Courville, and Y.~Bengio, ``Generative adversarial networks,'' 2014.

\bibitem{gan1}
A.~Shrivastava, T.~Pfister, O.~Tuzel, J.~Susskind, W.~Wang, and R.~Webb,
  ``Learning from simulated and unsupervised images through adversarial
  training,'' in {\em Proceedings of the IEEE conference on computer vision and
  pattern recognition}, pp.~2107--2116, 2017.

\bibitem{dietrich2021synthetic}
R.~Dietrich-Sussner, A.~Davari, T.~Scehaus, M.~Braun, V.~Christlein, A.~Maier,
  and C.~Riess, ``Synthetic glacier sar image generation from arbitrary masks
  using pix2pix algorithm,'' in {\em 2021 IEEE International Geoscience and
  Remote Sensing Symposium IGARSS}, pp.~4548--4551, IEEE, 2021.

\bibitem{rw_spatial1}
L.~{He}, J.~{Li}, C.~{Liu}, and S.~{Li}, ``Recent advances on
  spectral–spatial hyperspectral image classification: An overview and new
  guidelines,'' {\em IEEE Transactions on Geoscience and Remote Sensing},
  vol.~56, no.~3, pp.~1579--1597, 2018.

\bibitem{rw_spatial2}
P.~{Ghamisi}, E.~{Maggiori}, S.~{Li}, R.~{Souza}, Y.~{Tarablaka}, G.~{Moser},
  A.~{De Giorgi}, L.~{Fang}, Y.~{Chen}, M.~{Chi}, S.~B. {Serpico}, and J.~A.
  {Benediktsson}, ``New frontiers in spectral-spatial hyperspectral image
  classification: The latest advances based on mathematical morphology, markov
  random fields, segmentation, sparse representation, and deep learning,'' {\em
  IEEE Geoscience and Remote Sensing Magazine}, vol.~6, no.~3, pp.~10--43,
  2018.

\bibitem{chen2014deep}
Y.~Chen, Z.~Lin, X.~Zhao, G.~Wang, and Y.~Gu, ``Deep learning-based
  classification of hyperspectral data,'' {\em IEEE Journal of Selected topics
  in applied earth observations and remote sensing}, vol.~7, no.~6,
  pp.~2094--2107, 2014.

\bibitem{deep1}
K.~{Makantasis}, K.~{Karantzalos}, A.~{Doulamis}, and N.~{Doulamis}, ``Deep
  supervised learning for hyperspectral data classification through
  convolutional neural networks,'' in {\em 2015 IEEE International Geoscience
  and Remote Sensing Symposium (IGARSS)}, pp.~4959--4962, 2015.

\bibitem{deep2}
A.~Ben~Hamida, A.~Benoit, P.~Lambert, and C.~Ben-Amar, ``{DEEP LEARNING
  APPROACH FOR REMOTE SENSING IMAGE ANALYSIS},'' in {\em {Big Data from Space
  (BiDS'16)}} (S.~P. M.~P. Giorgio, ed.), (Santa Cruz de Tenerife, Spain),
  p.~133, {Publications Office of the European Union}, Mar. 2016.

\bibitem{deep3}
Y.~{Luo}, J.~{Zou}, C.~{Yao}, X.~{Zhao}, T.~{Li}, and G.~{Bai}, ``Hsi-cnn: A
  novel convolution neural network for hyperspectral image,'' in {\em 2018
  International Conference on Audio, Language and Image Processing (ICALIP)},
  pp.~464--469, 2018.

\bibitem{deep_lr}
N.~Audebert, B.~L. Saux, and S.~Lef{\`{e}}vre, ``Deep learning for
  classification of hyperspectral data: {A} comparative review,'' {\em CoRR},
  vol.~abs/1904.10674, 2019.

\bibitem{chen2017hyperspectral}
Y.~Chen, L.~Zhu, P.~Ghamisi, X.~Jia, G.~Li, and L.~Tang, ``Hyperspectral images
  classification with gabor filtering and convolutional neural network,'' {\em
  IEEE Geoscience and Remote Sensing Letters}, vol.~14, no.~12, pp.~2355--2359,
  2017.

\bibitem{aptoula2016deep}
E.~Aptoula, M.~C. Ozdemir, and B.~Yanikoglu, ``Deep learning with attribute
  profiles for hyperspectral image classification,'' {\em IEEE Geoscience and
  Remote Sensing Letters}, vol.~13, no.~12, pp.~1970--1974, 2016.

\bibitem{li2018data}
W.~Li, C.~Chen, M.~Zhang, H.~Li, and Q.~Du, ``Data augmentation for
  hyperspectral image classification with deep cnn,'' {\em IEEE Geoscience and
  Remote Sensing Letters}, vol.~16, no.~4, pp.~593--597, 2018.

\bibitem{he2020transferring}
X.~He and Y.~Chen, ``Transferring cnn ensemble for hyperspectral image
  classification,'' {\em IEEE Geoscience and Remote Sensing Letters}, vol.~18,
  no.~5, pp.~876--880, 2020.

\bibitem{cao2020hyperspectral}
X.~Cao, J.~Yao, Z.~Xu, and D.~Meng, ``Hyperspectral image classification with
  convolutional neural network and active learning,'' {\em IEEE Transactions on
  Geoscience and Remote Sensing}, vol.~58, no.~7, pp.~4604--4616, 2020.

\bibitem{b1}
W.~L. Buntine and A.~Weigend, ``Bayesian back-propagation,'' {\em Complex
  Syst.}, vol.~5, 1991.

\bibitem{blundell}
C.~Blundell, J.~Cornebise, K.~Kavukcuoglu, and D.~Wierstra, ``Weight
  uncertainty in neural networks,'' 2015.

\bibitem{b4}
J.~Denker and Y.~Lecun, ``Transforming neural-net output levels to probability
  distributions,'' in {\em Advances in Neural Information Processing Systems
  (NIPS 1990), Denver, CO, April 1991} (R.~Lippmann, J.~Moody, and
  D.~Touretzky, eds.), vol.~3, Morgan Kaufmann, 1991.

\bibitem{drop}
N.~Srivastava, G.~Hinton, A.~Krizhevsky, I.~Sutskever, and R.~Salakhutdinov,
  ``Dropout: A simple way to prevent neural networks from overfitting,'' {\em
  Journal of Machine Learning Research}, vol.~15, no.~56, pp.~1929--1958, 2014.

\bibitem{gdrop}
S.~Wang and C.~Manning, ``Fast dropout training,'' in {\em Proceedings of the
  30th International Conference on Machine Learning} (S.~Dasgupta and
  D.~McAllester, eds.), vol.~28 of {\em Proceedings of Machine Learning
  Research}, (Atlanta, Georgia, USA), pp.~118--126, PMLR, 17--19 Jun 2013.

\bibitem{bcnn_gubin}
Y.~Gal and Z.~Ghahramani, ``Bayesian convolutional neural networks with
  bernoulli approximate variational inference,'' 2016.

\bibitem{bayes_by_backprop1}
A.~Graves, ``Practical variational inference for neural networks,'' in {\em
  NIPS}, 2011.

\bibitem{softplus}
K.~Shridhar, F.~Laumann, and M.~Liwicki, ``Uncertainty estimations by softplus
  normalization in bayesian convolutional neural networks with variational
  inference,'' 2019.

\bibitem{mnist}
Y.~{Lecun}, L.~{Bottou}, Y.~{Bengio}, and P.~{Haffner}, ``Gradient-based
  learning applied to document recognition,'' {\em Proceedings of the IEEE},
  vol.~86, no.~11, pp.~2278--2324, 1998.

\bibitem{shridhar1}
K.~Shridhar, F.~Laumann, and M.~Liwicki, ``A comprehensive guide to bayesian
  convolutional neural network with variational inference,'' 2019.

\bibitem{kullback}
S.~Kullback and R.~A. Leibler, ``On information and sufficiency,'' {\em Ann.
  Math. Statist.}, vol.~22, no.~1, pp.~79--86, 1951.

\bibitem{variational}
D.~P. Kingma, T.~Salimans, and M.~Welling, ``Variational dropout and the local
  reparameterization trick,'' 2015.

\bibitem{aleatoric_or_epistemic}
A.~Der~Kiureghian and O.~Ditlevsen, ``Aleatory or epistemic? does it matter?,''
  {\em Structural Safety}, vol.~31, pp.~105--112, 03 2009.

\bibitem{dalla2010extended}
M.~Dalla~Mura, J.~Atli~Benediktsson, B.~Waske, and L.~Bruzzone, ``Extended
  profiles with morphological attribute filters for the analysis of
  hyperspectral data,'' {\em International Journal of Remote Sensing}, vol.~31,
  no.~22, pp.~5975--5991, 2010.

\bibitem{rf_breiman}
L.~Breiman, ``Random forests,'' {\em Machine Learning}, vol.~45, no.~1,
  pp.~5--32, 2001.

\end{thebibliography}

\newpage

\title{Supplementary material for "Bayesian Convolutional Neural Networks for Limited Data Hyperspectral Remote Sensing Image Classification"}

\twocolumn
	
\pagenumbering{arabic}
\maketitle

\section{Class-wise Performance}
\vspace{-17mm}
The class-wise performance of the Bayesian CNN classifiers on the Botswana, Pavia Centre, and Salinas datasets are presented in Tab.~\ref{tab:class_acc_botswana}, Tab.~\ref{tab:class_acc_pavia}, and Tab.~\ref{tab:class_acc_salinas}, respectively.

\begin{table}[H]
	\scriptsize
	\centering
	\caption{Bayesian network's class-wise accuracy on Botswana dataset.}
	\begin{tabular}{ @{}l@{ } c@{ } c@{ } c@{ } c@{}}
		\toprule
		Name                 & Train/Test    & RF                & CNN                & BNN                 \\
		\hline                                                                        
		Water                & 20/105        & 100.00 $\pm$ 0.00 &  99.88 $\pm$ 0.29  &  99.52 $ \pm $ 1.09 \\
		Hippo grass          & 20/41         & 88.66 $\pm$ 3.39  &  97.07 $\pm$ 5.15  &  98.78 $ \pm $ 3.32 \\			
		Floodplain grasses1  & 20/96         & 94.87 $\pm$ 3.94  &  92.67 $\pm$ 5.53  &  96.98 $ \pm $ 3.15 \\
		Floodplain grasses2  & 20/78         & 81.58 $\pm$ 6.67  &  88.93 $\pm$ 4.95  &  92.14 $ \pm $ 9.37 \\
		Reeds1               & 20/105        & 76.36 $\pm$ 4.99  &  80.68 $\pm$ 7.40  &  89.08 $ \pm $ 7.87 \\
		Riparian             & 20/105        & 63.00 $\pm$ 10.36 &  76.72 $\pm$ 7.86  & 73.48 $ \pm $ 10.62 \\		
		Firescare2           & 20/100        & 97.83 $\pm$ 1.10  &  97.25 $\pm$ 4.54  &  98.54 $ \pm $ 3.79 \\				
		Island interior      & 20/72         & 100.00 $\pm$ 0.00 &  85.00 $\pm$ 14.03 &  96.41 $ \pm $ 7.69 \\
		Acacia woodlands     & 20/127        & 90.65 $\pm$ 9.11  &  88.23 $\pm$ 8.23  &  93.95 $ \pm $ 6.32 \\	
		Acacia shrublands    & 20/94         & 73.03 $\pm$ 12.38 &  98.07 $\pm$ 4.01  &  95.83 $ \pm $ 5.20 \\
		Acacia grasslands    & 20/123        & 69.44 $\pm$ 14.83 &  85.28 $\pm$ 8.06  &  93.85 $ \pm $ 4.79 \\
		Short mopane         & 20/61         & 93.95 $\pm$ 2.31  &  98.77 $\pm$ 1.70  &  98.09 $ \pm $ 2.36 \\
		Mixed mopane         & 20/104        & 90.44 $\pm$ 7.07  &  99.19 $\pm$ 1.65  &  97.86 $ \pm $ 3.28 \\	
		Exposed soils        & 20/38         & 70.53 $\pm$ 8.91  &  85.53 $\pm$ 4.67  &  91.58 $ \pm $ 6.09 \\
		\hline                                                                        
		\multicolumn{2}{c}{Kappa}            & 0.8376$\pm$0.0277 & 0.8987 $\pm$0.0164 & 0.9315$ \pm$ 0.0128 \\
		\multicolumn{2}{c}{Overall Accuracy} & 85.02 $\pm$ 2.55  & 90.65 $\pm$ 1.52   &  93.68 $\pm$ 1.18   \\
		\multicolumn{2}{c}{Average Accuracy} & 79.36 $\pm$ 5.67  & 90.95 $\pm$ 5.58   &  94.01 $\pm$ 5.35   \\
		\hline
	\end{tabular}
	\label{tab:class_acc_botswana}
\end{table}

\begin{table}[H]
	\scriptsize
	\centering
	\caption{Bayesian network's class-wise accuracy on Pavia dataset.}
	\begin{tabular}{ @{}l@{ } c@{ }  c@{ } c@{ } c@{}}
		\toprule
		Name                 & Train/Test    & RF                & CNN               & BNN                 \\
		\hline                                                                                             
		Water                &  20/32976     & 98.98 $\pm$ 0.19  & 99.65 $\pm$ 0.36  &	99.10 $\pm$ 1.10   \\
		Trees                &  20/3789      & 66.27 $\pm$ 7.08  & 76.00 $\pm$ 6.83  &  74.88 $\pm$ 6.70   \\
		Asphalt              & 20/1535	     & 32.73 $\pm$ 2.05  & 36.93 $\pm$ 8.77  &  34.48 $\pm$ 4.24   \\
		Self-Blocking Bricks &  20/1333      & 79.02 $\pm$ 8.70  & 92.00 $\pm$ 12.48 &  92.80 $\pm$ 5.25   \\
		Bitumen              &  20/3282      & 49.98 $\pm$ 10.80 & 71.56 $\pm$ 17.79 &  65.61 $\pm$ 12.44  \\
		Tiles                &  20/4614	     & 91.52 $\pm$ 5.50  & 91.19 $\pm$ 6.07  &  86.63 $\pm$ 8.03   \\
		Shadows              &  20/3634	     & 74.78 $\pm$ 7.55  & 72.68 $\pm$ 10.77 &  80.02 $\pm$ 6.72   \\
		Meadows              &  20/21403     & 73.69 $\pm$ 12.63 & 92.67 $\pm$ 5.68  &  94.44 $\pm$ 4.57   \\
		Bare Soil            &  20/1422	     & 93.87 $\pm$ 2.06  & 97.67 $\pm$ 2.47  &  92.74 $\pm$ 7.51   \\
		\hline                                                                                             
		\multicolumn{2}{c}{Kappa}	         & 0.7730$\pm$0.0535 & 0.8849$\pm$0.0300 & 0.8943 $\pm$ 0.0220 \\
		\multicolumn{2}{c}{Overall Accuracy} & 83.64 $\pm$ 4.02  & 91.84 $\pm$ 3.00  & 92.52 $\pm$ 1.65    \\
		\multicolumn{2}{c}{Average Accuracy} & 66.08 $\pm$ 5.65  & 90.95 $\pm$ 5.58  & 94.01 $\pm$ 5.35    \\
		\hline
	\end{tabular}
	\label{tab:class_acc_pavia}
\end{table}

\begin{table}[H]
	\scriptsize
	\centering
	\caption{Bayesian network's class-wise accuracy on Salinas dataset.}
	\begin{tabular}{ @{}l@{ } c@{ } c@{ } c@{ } c@{}}
		\toprule
		Name                    & Train/Test & RF                & CNN               & BNN                 \\
		\hline                                                                       
		Broccoli green weeds 1    & 20/975   & 97.36 $\pm$ 2.53  & 96.21 $\pm$ 3.25  & 86.37 $\pm$ 22.95   \\
		Broccoli green weeds 2    & 20/1853  & 92.90 $\pm$ 5.91  & 90.75 $\pm$ 7.51  & 90.00 $\pm$ 5.90    \\
		Fallow                    & 20/958   & 57.88 $\pm$ 21.66 & 73.51 $\pm$ 17.83 & 71.55 $\pm$ 12.02   \\
		Fallow rough plow         & 20/667   & 99.59 $\pm$ 0.23  & 98.05 $\pm$ 1.59  & 99.00 $\pm$ 0.51    \\
		Fallow smooth             & 20/1309  &  94.74 $\pm$ 4.49 & 88.75 $\pm$ 7.41  & 87.68 $\pm$ 6.46    \\
		Stubble                   & 20/1950  & 97.47 $\pm$ 1.75  & 99.90 $\pm$ 0.28  & 99.64 $\pm$ 0.83    \\
		Celery                    & 20/1760  & 99.99 $\pm$ 0.03  & 98.69 $\pm$ 1.69  & 99.08 $\pm$ 0.80    \\
		Grapes untrained          & 20/5606  & 49.79 $\pm$ 19.51 & 58.32 $\pm$ 12.70 & 60.75 $\pm$ 11.42   \\
		Soil vineyard develop     & 20/3072  & 94.20 $\pm$ 5.03  & 95.65 $\pm$ 3.32  & 96.12 $\pm$ 2.38    \\
		Corn green weeds          & 20/1609  & 53.97 $\pm$ 17.86 & 54.29 $\pm$ 16.34 & 59.85 $\pm$ 18.65   \\
		Lettuce romaine 4wk       & 20/504   & 92.71 $\pm$ 3.36  & 86.74 $\pm$ 8.00  & 88.95 $\pm$ 7.38    \\
		Lettuce romaine 5wk       & 20/934   & 89.17 $\pm$ 15.41 & 97.61 $\pm$ 4.78  & 96.73 $\pm$ 3.24    \\
		Lettuce romaine 6wk       & 20/428   & 96.37 $\pm$ 3.51  & 97.06 $\pm$ 3.48  & 96.08 $\pm$ 5.33    \\
		Lettuce romaine 7wk       & 20/525   & 85.74 $\pm$ 9.66  & 95.04 $\pm$ 4.51  & 95.63 $\pm$ 3.65    \\
		Vineyard untrained        & 20/3604  & 63.42 $\pm$ 19.78 & 70.65 $\pm$ 17.38 & 67.17 $\pm$ 17.80   \\
		Vineyard vertical trellis & 20/874   & 99.32 $\pm$ 1.41  & 85.01 $\pm$ 9.24  & 91.33 $\pm$ 6.02    \\
		\hline                                                                       
		\multicolumn{2}{c}{Kappa}	         & 0.7481$\pm$0.0318 & 0.7857$\pm$0.0197 & 0.7870 $\pm$ 0.0278 \\
		\multicolumn{2}{c}{Overall Accuracy} & 77.23 $\pm$ 2.94  & 80.82 $\pm$ 1.74  & 80.77 $\pm$ 2.46    \\
		\multicolumn{2}{c}{Average Accuracy} & 80.27 $\pm$ 7.77  & 86.64 $\pm$ 7.46  & 86.62 $\pm$ 7.83    \\
		\hline
	\end{tabular}
	\label{tab:class_acc_salinas}
\end{table}

\begin{figure*}[t]
	\begin{minipage}[b]{0.32\linewidth}
		\centering
		\centerline{\includegraphics[width=1\linewidth]{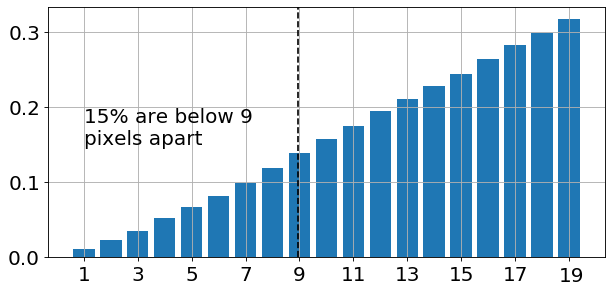}}
		\centerline{}\medskip
	\end{minipage}
	\begin{minipage}[b]{0.32\linewidth}
		\centering
		\centerline{\includegraphics[width=1\linewidth]{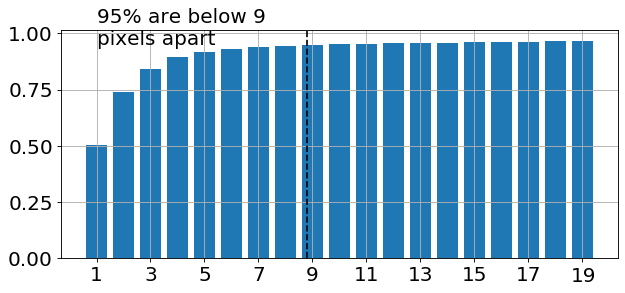}}
		\centerline{}\medskip
	\end{minipage}
	\begin{minipage}[b]{0.32\linewidth}
		\centering
		\centerline{\includegraphics[width=1\linewidth]{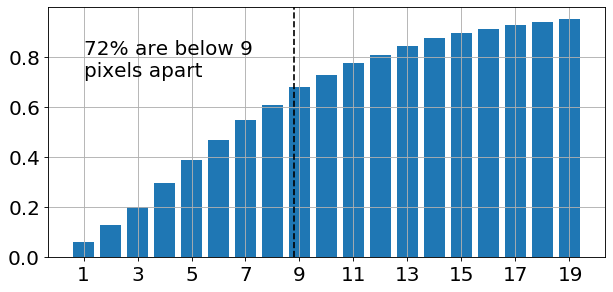}}
		\centerline{}\medskip
	\end{minipage}
	
	\begin{minipage}[b]{0.32\linewidth}
		\centering
		\centerline{\includegraphics[width=1\linewidth]{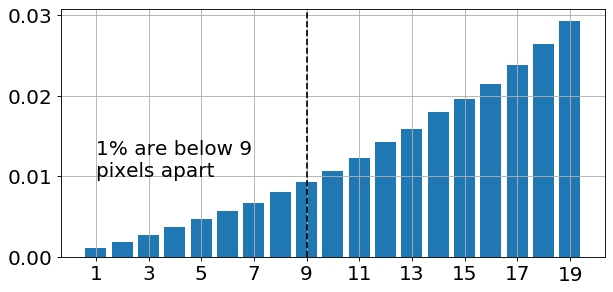}}
		\centerline{(a)}\medskip
	\end{minipage}
	\begin{minipage}[b]{0.32\linewidth}
		\centering
		\centerline{\includegraphics[width=1\linewidth]{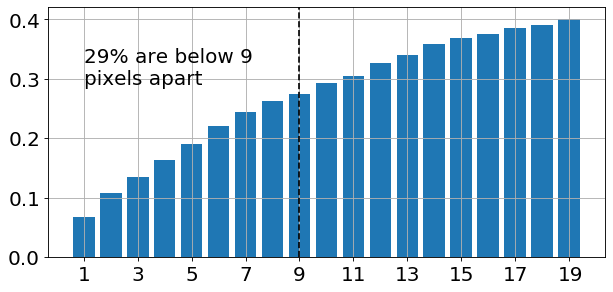}}
		\centerline{(b)}\medskip
	\end{minipage}
	\begin{minipage}[b]{0.32\linewidth}
		\centering
		\centerline{\includegraphics[width=1\linewidth]{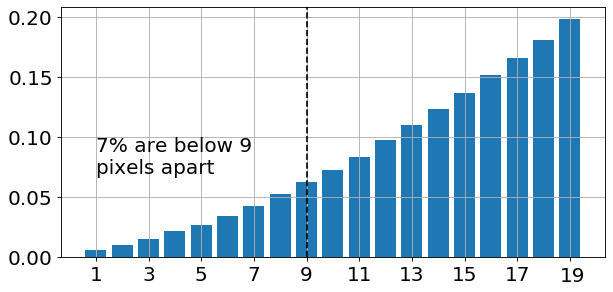}}
		\centerline{(c)}\medskip
	\end{minipage}
	
	\caption{Random (upper) and Connected-Component (lower) train split method comparison for (a) Pavia, (b) Botswana, and (c) Salinas datasets.}
	\label{fig:all_datasets_distance}
	
\end{figure*}

\section{Impact of the Train Split}

The training data has been selected as a connected component (CC) around a random pixel instead of randomly picking pixels throughout the image. This way, we minimize the chance of overlap between the training set, and the validation and test sets. To illustrate the difference caused by the random and CC sampling strategies and visualize how much the overlap decreases using the CC sampling policy, we carried out a random train split and a CC split.
Then, the histogram of the distance of all pixels in the validation and test sets from the closest training point of the same class is measured and plotted in \ref{fig:all_datasets_distance} as an accumulative bar plot.
The plot 
compares the 
widely used random pixel sampling (top) to the connected-component training
split used in this work (bottom). Plots (a), (b), (c) operate on the Botswana,
Pavia, and Salinas datasets. On the $x$-axis, the distance between
training and test pixels is shown. On the $y$-axis, the cumulative
relative frequencies are shown that a training pixel is within $x$ pixels next
to a test pixel.  
We use the patch size of $9\times 9$ pixels in the neural networks. Hence, we
highlight the overlap percentage at the distance of $9$ pixels.
If the distance of the patch centers is below $9$ pixels, an overlap occurs with at least one training patch from the same class. The less these occurrences happen, the more realistic the validation results would be.

As we can see in Fig. \ref{fig:all_datasets_distance}, the difference of the overlap between the two methods is quite large; in the Salinas dataset, using random train data extraction, more than $72\%$ of the validation and test data are less than $9$ pixels apart and are at least partly fed into the network during training. On the other hand, this number is around $7\%$ in the case of connected component sampling method. For the Botswana dataset, a random split partially feeds at least $95\%$ of the validation and test patches into the network during training, whereas $29\%$ is fed when using connected component training data. The same story applies to the Pavia dataset, and the overlap rate is more than $15\%$ in random split, as opposed to $1\%$ on the other method. These numbers suggest that using the connected component method would result in a more realistic measure of the model's generalization capability.


\end{document}